\documentclass[runningheads]{llncs}

\usepackage[mobile]{eccv}

\usepackage{eccvabbrv}

\usepackage{graphicx}
\usepackage{xcolor}
\usepackage{colortbl}
\usepackage{booktabs}
\usepackage{amsmath}
\usepackage{makecell}
\usepackage{multirow}
\usepackage{wrapfig}
\usepackage{array}
\usepackage{pifont}
\usepackage{algorithm}
\usepackage{algorithmic}
\usepackage{titletoc}
\usepackage{marvosym} 
\usepackage[accsupp]{axessibility}

\definecolor{lightgray}{gray}{0.5}
\newcommand{\methodname}{\textcolor{black}{RosettaSim}}

\usepackage{hyperref}
\hypersetup{hypertexnames=false}

\usepackage{orcidlink}

\makeatletter
\let\orig@fnsymbol\@fnsymbol
\renewcommand{\@fnsymbol}[1]{\ifcase#1\or \textrm{\Letter}\else\orig@fnsymbol{#1}\fi}
\makeatother

\begin{document}

\title{Long-term Traffic Simulation via Structured Autoregressive Modeling}

\author{Lingyu Xiao\orcidlink{0009-0006-6500-7387}, Zexin Feng\orcidlink{0009-0007-5247-2922}, \and Xintao Yan\thanks{Corresponding author.}\orcidlink{0000-0002-0569-5628}}
\authorrunning{L.~Xiao et al.}
\institute{The University of Hong Kong, Pokfulam, Hong Kong \\
\email{\{lingyu.xiao,zexinfeng\}@connect.hku.hk} \
\email{xintaoy@hku.hk} \\
\email{\url{https://sephirex-x.github.io/rosettasim/}}
} 

\maketitle

\begin{abstract}

Interactive traffic simulation is a vital world model for autonomous driving.
A central challenge in long-horizon simulation is modeling sustained multi-agent interactions, which is further exacerbated by dynamic token cardinality as agents continuously enter and exit the scene.
In this work, we propose that the solution lies in the synergy between the architectural inductive biases and statistical priors of large-scale sequence models, e.g., Large Language Models (LLMs).
Our probing experiments reveal that the transferability of attention mechanisms and the distributional consistency between motion tokens and natural language enable small-scale, heavily frozen LLMs to rapidly adapt to traffic modeling.
Building on this insight, we introduce \methodname{}, a unified framework that projects scene topology, agent states, and spawning intents into a structured autoregressive stream with variable length, achieving both strong short-term accuracy and stable long-horizon simulation fidelity.
Furthermore, evaluating extended rollouts presents yet another hurdle, as one-to-one agent correspondence inevitably fades over time. To address this, we introduce Retrieval-based Traffic Evaluation (RTE), which retrieves semantically similar real-world scenarios as context-aware reference anchors.
Experiments on the Waymo Open Sim Agent Challenge (WOSAC) demonstrate that \methodname{} achieves state-of-the-art performance in both short- and long-term simulation. Furthermore, RTE exhibits a stronger correlation with standard metrics ($r=0.83$) than existing approaches ($r=0.74$), indicating improved alignment with long-horizon simulation fidelity.
\end{abstract}

\section{Introduction}
\label{sec:intro}
\begin{figure*}
    
    \includegraphics[width=1\textwidth]{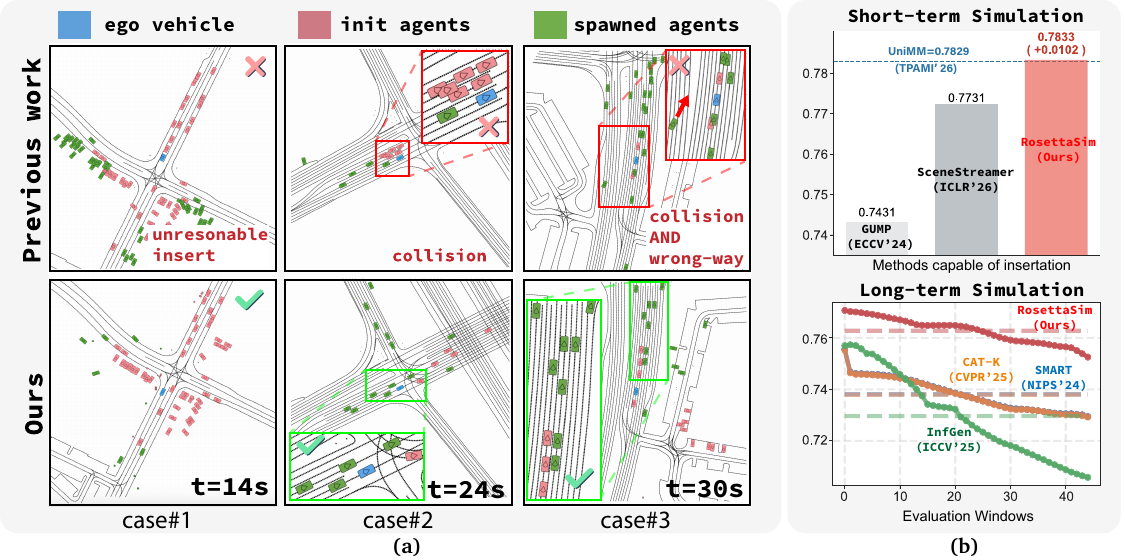}
    \vspace{-8mm}
    \caption{\textbf{Performance.} (a) Visualization of long-term simulation compared with~\cite{infgen_iccv}. Mere demo videos can be found on the webpage. (b) Quantitative performance comparison on WOMD under close-loop short/long-term simulation.}
    \label{fig:teaser}
    \vspace{-8mm}
\end{figure*}
Traffic simulation underpins the scalable and safe development of autonomous driving systems~\cite{yan2023learning, feng2023dense, liu2025autonomous, feng2026breaking}. While prior works~\cite{philion2023trajeglish,nips24smart,zhang2024catk,wang2025llm,zhou2024behaviorgpt,rowe2024ctrl,yan2025distributionally} excel under short-term setting (typically $<8$s), they struggle in long-term, trip-level simulation where agents continuously enter and exit the scene.
To address this, recent works~\cite{Tan_2025_CVPR,infgen_iccv} have reframed long-term traffic simulation as a joint modeling problem that integrates scene generation~\cite{rowe2025scenario,tan_2021_scenegen,trafficgen,chitta2024sledge} (agent injection) and motion generation within a unified rollout process. Despite this progress, two fundamental challenges remain unresolved: (i) a persistent performance trade-off when jointly modeling agent population dynamics and interaction-aware motion, and (ii) the lack of reliable evaluation protocols for long-term simulation.

The first challenge arises from the inherently different learning objectives of scene and motion generation. Scene generation focuses on spatial distributions (predicting \textit{where} and \textit{when} a new agent appears), whereas motion generation emphasizes temporal consistency (predicting \textit{how} agents move). Existing approaches~\cite{infgen_iccv,peng2025infgen} that jointly optimize these objectives often struggle to balance the two tasks.
As shown on Fig.~\ref{fig:teaser}(b) upper, the most recent work SceneStreamer~\cite{peng2025infgen} falls large behind against specialized short-term models (e.g., UniMM~\cite{lin2025revisit}).
This imbalance suggests that current modeling paradigms cannot adequately represent both agent population dynamics and long-term motion dependencies simultaneously.
In parallel, large-scale sequence models, e.g., Large Language Models (LLMs), have demonstrated strong capabilities in modeling long-range dependencies across both homogeneous sequences~\cite{brown2020language,touvron2023llama,yang2025qwen3,grattafiori2024llama} (e.g., text) and heterogeneous modalities~\cite{lu2022frozen,liu2023visual,kim2024openvla} (e.g., vision, robotics control). However, their application to traffic simulation remains limited.
Current approach typically uses language models as a conditioning tool for controllable simulation~\cite{tan2024promptable} or scene generation~\cite{hu2024gump}. Other efforts redesign components inspired by language models~\cite{zhou2024behaviorgpt,wang2025llm,peng2025infgen} or simply draw analogies to next-token prediction~\cite{philion2023trajeglish,nips24smart,infgen_iccv}. None of these works provide justification for whether LLMs' modeling capabilities for structural sequence can be applied to traffic simulation, let alone long-horizon simulation. Therefore, an intriguing question arises: Can large-scale sequence models effectively model traffic dynamics without relying on linguistic supervision, and if so, why?

We posit that the potential of {LLMs} lies not in linguistic semantics, but in their ability to model structured, compositional sequences autoregressively.  Specifically, as shown in Fig.~\ref{fig:zipfian}, we observe that the frequency distribution of discretized traffic motion tokens
closely mirrors the {statistical properties} of natural language. {Motivated by~\cite{chan2022data}, we hypothesize that sequence models trained under similar regimes provide a highly optimized initialization manifold for traffic dynamics.} To empirically validate this, we conducted probing experiments on motion generation using fully and partially frozen LLMs. The rapid adaptation of even a small-scale, heavily frozen LLM confirms that these models effectively process traffic dynamics as structured sequential data with long-range dependencies, akin to learning a ``foreign language''.
{Furthermore, we observe that LLM exhibits highly similar attention locality~\footnote{Attention locality is defined as the sum of attention weights along the main diagonal and its immediately adjacent diagonals, measuring the proportion of attention a token allocates to itself and its direct neighbors.} when processing language and motion data separately, further substantiating our hypothesis.}
Building on these insights, we propose a unified paradigm that seamlessly projects geometric topology, existing agent states, and generation intents into a structured autoregressive token sequence with variable length.
{We name our method \methodname{}, acting like a ``Rosetta Stone'' between language and motion}.
By {modeling the sequence structurally}, \methodname{} can inherit both the modeling capability under flexible token length and the structural prior from pretraining, thus significantly mitigating the performance trade-off between scene and motion generation.
As shown in Fig.~\ref{fig:teaser}(a), \methodname{}
demonstrates improved traffic flow realism and multi-agent interaction consistency compared to the prior method~\cite{infgen_iccv}.

The second bottleneck lies in evaluation. Standard metrics (e.g., WOSAC~\cite{montali2023waymo}) rely on strict one-to-one agent matching, making them inapplicable for dynamic long-term rollouts. Existing  workarounds~\cite{infgen_iccv,Tan_2025_CVPR} compare rollout histograms against entire validation sets; however, because the aggregated distribution of the entire validation set is not context-aware, global matching yields misleadingly high scores for trivial policies. For example, it is unreasonable to ask a high-speed scenario to match the overall distribution, which is dominated by low-speed scenarios. To resolve this, we propose the Retrieval-based Traffic Evaluation (RTE) framework. RTE mitigates statistical bias by leveraging a pretrained VAE~\cite{rowe2025scenario,kingma2013auto} latent space to retrieve the top-K most semantically similar real-world scenarios. These serve as grounded reference anchors, ensuring a rigorous, context-aware, and fair evaluation.

Though extensive close-loop experiments on Waymo Open Motion Dataset (WOMD)~\cite{ettinger2021womd}, we demonstrate that our method achieves state-of-the-art performance in both short-term benchmarking~\cite{montali2023waymo} and long-term simulation compared with its counterparts (see Fig~\ref{fig:teaser} (b)).

In summary, our contributions are threefold:

\textbf{-} We introduce a unified paradigm for long-horizon traffic simulation that formulates multi-agent scene evolution as a conditional sequence generation problem with dynamically varying token cardinality, enabling consistent alignment between agent population dynamics and interaction-aware motion generation.

\textbf{-} We propose Retrieval-based Traffic Evaluation (RTE), a context-aware evaluation framework that retrieves semantically similar real-world scenarios as reference anchors for metric computation, providing a more reliable assessment of long-horizon traffic simulation.

\textbf{-} We present extensive empirical analysis demonstrating why and how structural priors from large-scale sequence models (LLMs) can be adapted for interactive traffic simulation, highlighting their effectiveness in modeling multi-agent dynamics beyond language domains.

\section{Related Works}
\textbf{Closed-loop Traffic Simulation.} While early works adapted motion forecasting models~\cite{montali2023waymo,shi2022motion}, recent state-of-the-art formulates simulation as next-token prediction~\cite{nips24smart,philion2023trajeglish,zhang2024catk,zhao2024kigras}, often enhanced by reinforcement learning~\cite{guo2025decompgail,pei2025advancing} or advanced tokenization~\cite{zhangtrajtok}. However, these approaches primarily target short-term generation and fail to address the complex population dynamics in long-term simulations.

\noindent\textbf{Language Models for Non-Language Tasks.} Beyond NLP, LLMs are increasingly adapted for diverse domains via task-to-text formatting~\cite{dinh2022lift,mirchandani2023large,tan2024promptable} or instruction tuning~\cite{kim2024openvla,liu2023visual}. Crucially, fully frozen LLMs have proven effective as structural priors for continuous tasks like 3D scene understanding~\cite{lu2022frozen,pangfrozen}. Yet, exploiting these pretrained priors to orchestrate autoregressive long-term traffic dynamics remains a critical open gap that our work addresses.

\section{Preliminary}

We formulate the long-term traffic simulation problem as a conditional sequence generation task~\cite{seff2023motionlm,philion2023trajeglish}. Let $s^i_t$ denote the state token (e.g., position, heading, velocity) of the $i$-th agent at timestep $t$. The global traffic scene is represented as $\mathcal{S}_t = \{s^1_t, s^2_t, \dots, s^{N_t}_t\}$, where the active agent count $N_t$ varies dynamically over an extended simulation horizon $T$~\cite{infgen_iccv,Tan_2025_CVPR}. Unlike standard short-term simulations that assume a fixed set of agents, long-term simulations must continuously manage agent entries and exits. Therefore, we decompose the standard transition probability $p(\mathcal{S}_{t+1} \mid \mathcal{S}_t, \mathcal{C})$~\cite{nips24smart,philion2023trajeglish} into two tractable processes: \textit{Parallel Motion Update} and \textit{Autoregressive Agent Generation}.\subsection{Formulation}\label{sec:formulation}We introduce an intermediate state $\tilde{\mathcal{S}}_{t+1}$ representing the motion-updated scene of existing agents before any topological changes. The next-scene generation is factorized as:
\begin{equation}
    p(\mathcal{S}_{t+1} \mid \mathcal{S}_t, \mathcal{C}) = \underbrace{p_{\text{agent}}(\mathcal{S}_{t+1} \mid \tilde{\mathcal{S}}_{t+1}, \mathcal{C})}_{\text{Autoregressive Generation}} \cdot \underbrace{p_{\text{motion}}(\tilde{\mathcal{S}}_{t+1} \mid \mathcal{S}_t, \mathcal{C})}_{\text{Parallel Motion Update}}.
\end{equation}

For \textit{Parallel Motion Update},
 given the current scene $\mathcal{S}_t$, the future states of all $N_t$ existing agents are predicted simultaneously. To ensure simulation efficiency, we approximate this joint update by assuming conditional independence among agents given the rich historical scene context $\mathcal{S}_t$ and map $\mathcal{C}$:
\begin{equation}
p_{\text{motion}}(\tilde{\mathcal{S}}_{t+1} \mid \mathcal{S}_{t},\mathcal{C}) = \prod_{i=1}^{N_{t}} p(\tilde{s}_{t+1}^{i} \mid \mathcal{S}_{t},\mathcal{C}).
\end{equation}

For \textit{Autoregressive Agent Generation},
conditioned on the motion-updated intermediate scene $\tilde{\mathcal{S}}_{t+1}$, the model autoregressively determines the final new scene $\mathcal{S}_{t+1}$. This process dynamically handles the spawning of new agents, ensuring consistent traffic density and logical boundary interactions.
\subsection{Metrics for Traffic Simulation}

Standard short-term evaluations rely on WOSAC metrics~\cite{montali2023waymo} (kinematic, interactive, and map-based realism) computed via strict one-to-one agent correspondence with ground truth logs. For long-term settings where agents dynamically enter and exit, previous works~\cite{infgen_iccv,Tan_2025_CVPR} augment these with statistics like number of enter/exit, distance of enter/exit (\#Enter, \#Exit, $D_{enter}$, $D_{exit}$) and compute aggregated distribution divergences against the entire validation set.

\section{\methodname{}}

\subsection{Motivation}
\begin{figure}
    \vspace{-5mm}
    \centering
    \includegraphics[width=1\textwidth]{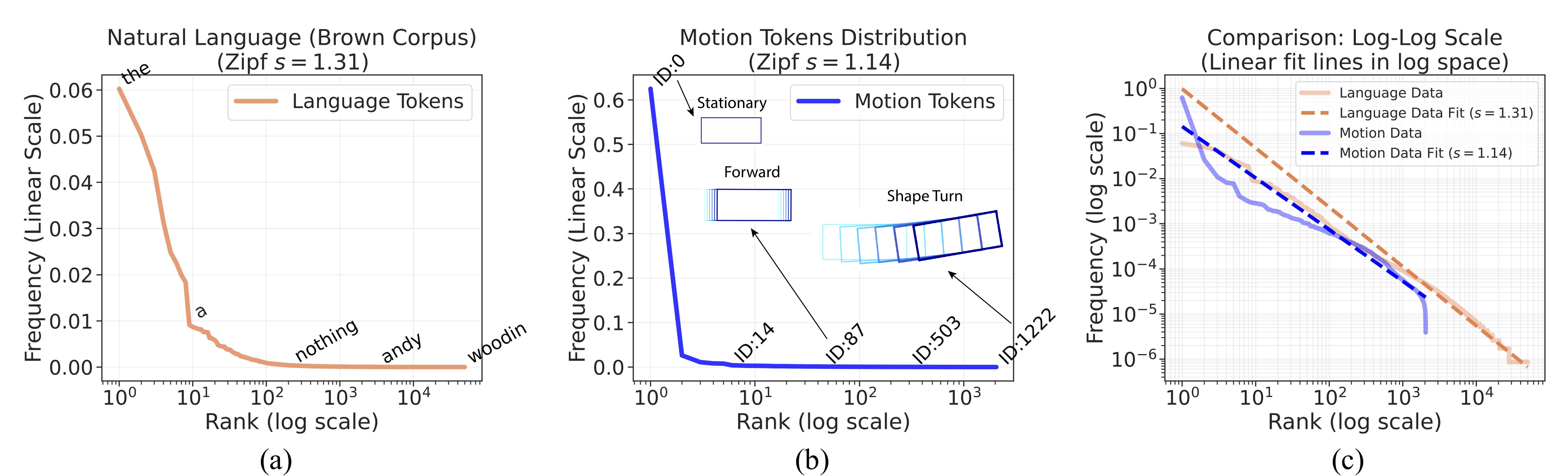}
    \vspace{-5mm}
    \caption{The token frequency distribution of traffic motion tokens on WOMD~\cite{montali2023waymo}. The distribution follows Zipf's law~\cite{ZipfGeorgeKingsley1965Hbat}, similar to natural language. (a) The frequency proportion of the language tokens~\cite{FrancisW.Nelson1979Bcmm}.(b) The frequency proportion of motion tokens. (c) The comparison between motion tokens and language tokens in log-log space.}
    \label{fig:zipfian}
    \vspace{-5mm}
\end{figure}
Recent advancements in traffic simulators~\cite{yan2023learning,nips24smart,philion2023trajeglish,zhang2024catk,wang2025llm} have increasingly embraced the formulation of large sequence models, adopting next-token-prediction paradigms and tokenizer-based designs~\cite{zhangtrajtok}. However, extending standard short-term traffic simulation to long-horizon simulation presents a significant challenge. It requires a unified autoregressive framework capable of simultaneously handling continuous motion generation and discrete agent population dynamics (i.e., agent entering and exiting)~\cite{peng2025infgen,infgen_iccv}. Optimizing such a complex, unified sequence model from scratch is notoriously difficult, often suffering from training instability and severe compounding errors over long horizons.

To address this optimization bottleneck, we step back to investigate the fundamental statistical properties of discretized traffic dynamics. Inspired by~\cite{chan2022data}, we empirically discover a striking alignment: the distribution of traffic motion tokens strictly follows Zipf's law~\cite{ZipfGeorgeKingsley1965Hbat} (Fig.~\ref{fig:zipfian}(c)), a hallmark statistical property of natural language. As shown in Fig.~\ref{fig:zipfian}(a) and (b), the frequency of motion tokens in WOMD~\cite{montali2023waymo} exhibits nearly identical distribution characteristics to language tokens, where a few specific actions (e.g., cruising, static) dominate the occurrences, while complex maneuvers (e.g., sharp turns) appear infrequently.

This critical observation suggests that pretrained LLMs may inherently encode similar token distributions during their massive text pretraining. Rather than treating traffic simulation as a completely isolated domain requiring from-scratch optimization, LLMs offer a highly optimized structural prior and initialization space. By conceptualizing traffic dynamics as a structured ``foreign language'', we can directly leverage these pretrained priors to stabilize the unified generation of motions and population dynamics.

\subsection{Architecture}
\begin{figure}[h]
    \vspace{-7mm}
    \centering
    \includegraphics[width=1\textwidth]{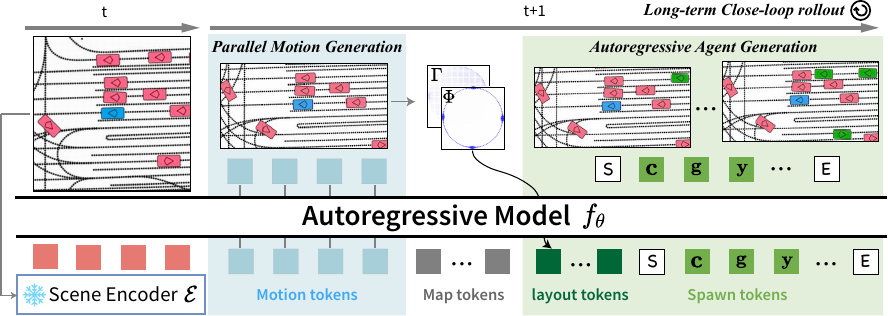}
    \caption{Architecture of \methodname{}. We formulate the long-term traffic simulation as a structured sequence generation problem.}
    \label{fig:architecture}
    \vspace{-10mm}
\end{figure}
\subsubsection{Parallel Motion Generation}
The primary goal is to effectively leverage the capabilities of pretrained large sequence model $f_\theta$, e.g., LLMs. The overall architecture is illustrated in Fig.~\ref{fig:architecture}. We treat the pretrained transformer layers on~\cite{nips24smart} as our scene encoder $\mathcal{E}$ that encodes the agent-centered scene context into agent tokens: $\mathbf{a}^{1:N_t}_t=\mathcal{E}(\mathcal{S}_t,\mathcal{C})$. The dimension is aligned with LLM by MLP. Then, we use a set of learnable motion tokens $\mathbf{mo}^{1:N_t}$ to retrieve the motion context from agent tokens via self-attention in the language model $f_\theta$:
\begin{equation}
    \mathbf{mo}^{1:N_t}_{t} = f_\theta(\mathbf{mo}^{1:N_t}, \mathbf{a}^{1:N_t}_t).
\end{equation}
To be noted that, before feeding all tokens into LLM, we will add agent-wise sinusoidal position embeddings to both agent tokens and motion tokens to encode the agent identity:
\begin{equation}
    \mathbf{mo}^i = \mathbf{mo}^i + \text{SinPE}(i), \quad \mathbf{a}^i_t = \mathbf{a}^i_t + \text{SinPE}(i),
\end{equation}
where $\text{SinPE}(\cdot)$ is the sinusoidal position embedding function.
The updated motion tokens $\mathbf{mo}^{1:N_t}_{t}$ are then projected to motion logits for each agent,

which are used to update motion in parallel:
\begin{equation}
    p_{\text{motion}}(\cdot \mid \mathcal{S}_{t},\mathcal{C}) = \text{Softmax}(\texttt{mlp}_{motion}(\mathbf{mo}^{1:N_t}_{t}))
\end{equation}

For those newly spawned agents, we use a learnable embedding for $\mathcal{E}$ to get $\mathbf{a}$, instead of using a separate head to predict its initiated speed.

\subsubsection{Autoregressive Agent Generation}
The process of agents entering the scene is analogous to sequence generation: each newly spawned agent depends on the existing traffic context, while the total number of agents remains variable but must remain semantically coherent.  Importantly, newly generated agents should not disrupt the motion trajectories of agents already present in the scene. From this perspective, the current scene and motion tokens, $\{\mathbf{a}^{1:N_t}, \mathbf{mo}^{1:N_t}\}$ can be interpreted as a contextual prompt that conditions how traffic evolves.

To improve the accuracy and stability of this generation process, after obtaining the intermediate state $\tilde{\mathcal{S}}_{t+1}$, we introduce two additional topological-aware tokens as contextual input, map tokens $\mathbf{m}^{1:M_t}$ and layout tokens $\mathbf{l}^{1:N_t}$. The map tokens are extracted from a pretrained map encoder within scene encoder $\mathcal{E}$ and encode fine-grained geometric and topological relationships of the road network: $\mathbf{m}^{1:M_t}=\mathcal{E}(\mathcal{C})$. The layout tokens are consisted to two parts, grid tokens $\mathbf{g}$ and yaw tokens $\mathbf{y}$. Specifically, we will decompose the intermediate state $\tilde{\mathcal{S}}_{t+1}$ into position and orientation under ego-centered coordination. Following~\cite{infgen_iccv}, we tokenize the position into a set of discretized grid index as well as the orientation into a set of discretized yaw index. Therefore, the layout tokens can be mathematically interpreted as:
\begin{equation}
    \mathbf{l}^i = \mathbf{g}^i + \mathbf{y}^i + \text{SinPE}(i),\quad \mathbf{g}^i = \texttt{emb}_g(\Gamma(\tilde{s}^i_{t+1})), \quad \mathbf{y}^i = \texttt{emb}_y(\Phi(\tilde{s}^i_{t+1})),
\end{equation}
where $\Gamma(\cdot)$ and $\Phi(\cdot)$ are tokenizers that map the grid index and yaw index from the intermediate state $\tilde{s}^i_{t+1}$, respectively. $\texttt{emb}_g(\cdot)$ and $\texttt{emb}_y(\cdot)$ are the embedding layers for grid and yaw tokens. Similar to motion tokens, we also add agent-wise sinusoidal position embeddings to layout tokens to encode agent identity.

As shown in Fig.~\ref{fig:architecture}, we use three tokens to represent the newly spawned agent: type token $\mathbf{c}$, grid token $\mathbf{g}$, and yaw token $\mathbf{y}$. During inference, we will first sample the agent type by agent type head: $c^{j} \sim \mathcal{P}(\texttt{mlp}_{type}(f_\theta(\texttt{<SOS>}+\text{SinPE}(j))) $. Here, $\texttt{<SOS>}$ is the special token that stands for the start of autoregressive generation and $\mathcal{P}(\cdot)$ denotes sampling from a categorical distribution parameterized by the normalized logits. Then, conditioned on the sampled type token $\mathbf{c}^{j}$, $f_\theta$ will generate the grid token $\mathbf{g}^{j}$ and yaw token $\mathbf{y}^{j}$ autoregressively:

\begin{equation}
    \resizebox{0.90\linewidth}{!}{
    $g^{j} \sim \mathcal{P}(\texttt{mlp}_{grid}(f_{\theta}(\mathbf{c}^{j} + \text{SinPE}(j)))), \quad y^{j} \sim \mathcal{P}(\texttt{mlp}_{yaw}(f_{\theta}(\mathbf{g}^{j} + \mathbf{c}^{j} + \text{SinPE}(j)))).$
    }
\end{equation}
$j$ is the index of the newly spawned agent. If the model decides to stop spawning new agents, it will sample a special token $\texttt{<EOS>}$ from $\texttt{mlp}_{type}$. $\mathbf{g}^{j}$ and $\mathbf{y}^{j}$ are embedded from the same embedding layers as layout tokens. Therefore, the final number of agents are $N_{t+1} = N_t + J$ and the final generated sequence is
$\mathbf{s}=[\texttt{<SOS>}, \mathbf{c}^1, \mathbf{g}^1, \mathbf{y}^1, \ldots, \mathbf{c}^J, \mathbf{g}^J, \mathbf{y}^J, \texttt{<EOS>}] \in \mathbb{R}^{(3J + 2) \times D}$.
$D$ is the dimension of $f_\theta$.

\subsection{Training}
Unlike previous works that either use complex weighted losses~\cite{infgen_iccv} or multiple stages~\cite{peng2025infgen}, our training paradigm is fully one-stage with simple cross-entropy losses.

Specifically, we optimize the following loss function:
\begin{align}
    \mathcal{L} &= \text{CausalCrossEntropy}(f_\theta(\texttt{seq}), \texttt{label}),  \\
    \nonumber \texttt{seq} &= [\mathbf{a}^{1:N_t},  \mathbf{mo}^{1:N_t}, \mathbf{m}^{1:M_t}, \mathbf{l}^{1:N_t}, \mathbf{s}], \\
    \nonumber \texttt{label} &= [\texttt{pad}^{1:N_t}, \hat{{mo}}^{1:N_t}, \texttt{pad}^{1:M_t+N_t}, \hat{{
    s}}].
\end{align}
$\hat{mo}^{1:N_t}$ and $\hat{s}$ are the ground-truth motion tokens and spawn tokens, respectively. $\texttt{pad}$ is the padding label that will be ignored during loss computation. {Detailed tokenization and training hyperparameters will be described in the Appendix.}

\section{Retrieval-based Long-term  Traffic Evaluation}
\label{sec:RTE}
\subsection{Motivation}

Evaluating long-term simulation is fundamentally challenging because dynamic agent entering/exiting invalidates strict one-to-one matching. Prior works~\cite{infgen_iccv,Tan_2025_CVPR} circumvent this by slicing rollouts into short windows and computing divergences against the entire validation set. However, this global matching suffers from severe statistical bias. For instance, forcing a high-speed scenario to match an overall validation distribution dominated by low-speed logs~\cite{xiao2024easychauffeur,xiao2025learning} yields penalizing and misleading evaluations, particularly on Kinematic metrics (empirically confirmed in Fig.~\ref{fig:correlation} and Sec.~\ref{sec:RTE_correlation}).

\subsection{Approach}

To address the aforementioned problems, one intuitive idea is to find the most similar scenario for the validation set for each simulated scenario and compute the distribution between them instead. Inspired by the LPIPS metric~\cite{zhang2018unreasonable} in image generation, we propose to measure the similarity between two driving scenarios via a pretrained traffic-specific model's latent space. Works like~\cite{ding2023realgen, ding2025realdrive} has demonstrated that the latent space from a well-trained model can reflect the semantic similarity between two driving scenarios. Therefore, given a segment of long-term policy rollout, we can retrieve the top-K most similar scenarios from the validation set as reference and compute the statistical metric between them.

\begin{figure}[h]
    \vspace{-3mm}
    \centering
    \includegraphics[width=1.0\textwidth]{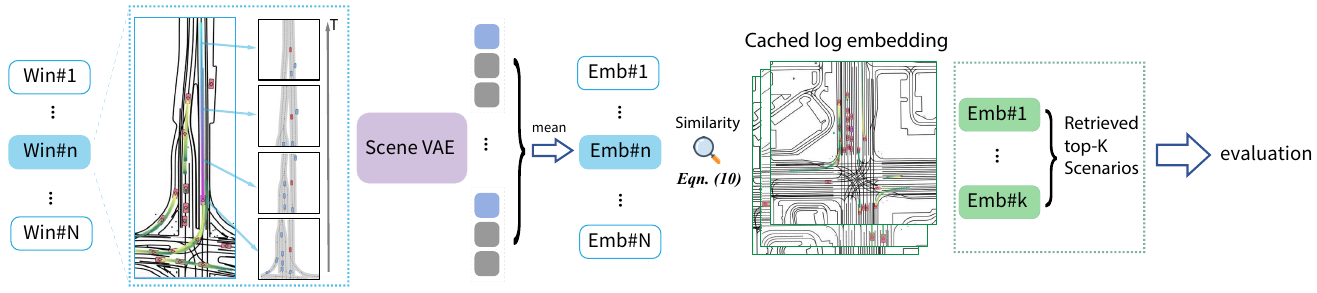}
    \caption{Overview of our proposed retrieval-based evaluation (RTE) framework for long-term traffic simulation.}
    \vspace{-3mm}
    \label{fig:RTE}
\end{figure}

The whole pipeline is shown in Fig.~\ref{fig:RTE}. Specifically, given a simulated rollout $\mathcal{S}_{0:T}$, we will first divide it into several short-term segments $\{\mathcal{S}_{t:t+\tau}\}$, where $\tau$ is the length of each segment. Then, we will encode each segment into a latent representation via a scene VAE~\cite{kingma2013auto} $\mathcal{E}_{\phi}$ that was pretrained for scenario generation. We directly use the official checkpoint from ~\cite{rowe2025scenario} without any fine-tuning to ensure reproducibility. The latent representation is formulated as:
\begin{equation}
    \mathbf{z} = \frac{1}{\tau} \sum_{t=1}^{\tau} \mathcal{E}_{\phi}(\mathcal{S}_{t}), \quad \mathbf{z}=[\mathbf{z}_\text{object}, \mathbf{z}_\text{map}].
    \label{eq:vae_encode}
\end{equation}
Where $\mathbf{z}_\text{object}:=(\mu_{o}, \sigma_{o})$ and $\mathbf{z}_\text{map}:=(\mu_{m}, \sigma_{m})$ are the object-level and map-level latent representation, respectively. We will also encode all the log segments in the validation set into the same latent space and build a retrieval database $\hat{\mathbf{z}}$.
During evaluation, for each simulated segment, we will retrieve the top-K similar segments from the validation set based on the Wasserstein distance~\cite{villani2009wasserstein} between their latent representations:
\begin{align}
    \text{similarity}(\mathbf{z}, \hat{\mathbf{z}}) &= W_2(\mathbf{z}_\text{object}, \hat{\mathbf{z}}_\text{object}) + \lambda \cdot W_2(\mathbf{z}_\text{map}, \hat{\mathbf{z}}_\text{map}).
    \label{eq:retrieval}
\end{align}

Finally, we will compute the evaluation metrics between the simulated segment and the retrieved references.

\subsection{Formulation of the Metrics}

To rigorously evaluate long-term simulation, we propose the Realism Meta Metric F1 (RMM-F1), formulated as the harmonic mean of two critical dimensions: 1) \textit {Behavior Realism}: Aggregates WOSAC-like metrics between generated segments and retrieved real-world references to capture fine-grained kinematic, interactive, and map-based fidelity. 2) \textit{Traffic Flow Realism}: Evaluates macro-level population dynamics using agent entry/exit counts against the global log distribution. We use the harmonic mean ($\beta=1$) to equally weight micro-level interactions and macro-level flows, analogous to the standard F1-score. This strictly penalizes models underperforming in either dimension, preventing deceptive compensation. Full formulations are detailed in the Appendix.

\section{Results}
\subsection{Short-term Traffic Simulation}
In this section, we manage to answer the following questions: 1) \textit{How does our method perform on short-term traffic simulation compared with other methods?} 2) \textit{How efficiently can large-scale sequence models adapt to continuous traffic dynamics utilizing their structural priors?}
\begin{table}[h]
    \vspace{-5mm}
\centering
\caption{Comparison of different methods on WOSAC 2025 \textit{test split}. `*' indicates methods from WOSAC 2024. \textbf{Bold} and \underline{underlined} denote the best and second-best performance. The method in \textcolor{gray}{gray} (CAT-K) utilizes closed-loop tuning and is listed for reference. `-' indicates the metric is not applicable.}
\label{tab:comparison}
\resizebox{\linewidth}{!}{
\begin{tabular}{l c c c c c c c}
\toprule
Method &
LLM Util. &
\begin{tabular}[c]{@{}c@{}}Agent \\ Insertion \end{tabular} &
\begin{tabular}[c]{@{}c@{}}Composite \\ Score $\uparrow$\end{tabular} &
\begin{tabular}[c]{@{}c@{}}Kinematic \\ metrics $\uparrow$\end{tabular} &
\begin{tabular}[c]{@{}c@{}}Interactive \\ metrics $\uparrow$\end{tabular} & \begin{tabular}[c]{@{}c@{}}Map-based \\ metrics $\uparrow$\end{tabular} &
\begin{tabular}[c]{@{}c@{}}min \\ ADE $\downarrow$\end{tabular} \\
\midrule
ProSim*~\cite{tan2024promptable}      & Llama2-7B~\cite{touvron2023llama} & \ding{55} & 0.7180 & 0.4010 & 0.7780 & 0.8854 & -  \\
LLM2AD~\cite{wang2025llm}  & Arch. Inspired& \ding{55}   & 0.7779 & 0.4846 & 0.8048 & 0.9109 & \textbf{1.2827}  \\
SMART~\cite{nips24smart}    & - & \ding{55}     & 0.7814 & 0.4854 & 0.8089 & 0.9153 & 1.3931 \\
SimFormer~\cite{wang2025improving} & - & \ding{55}   & 0.7820 & 0.4920 & 0.8060 & \textbf{0.9167} & 1.3221  \\
UniMM~\cite{lin2025revisit}   & - & \ding{55}     & \underline{0.7829} & \textbf{0.4914} & \underline{0.8089} & \underline{0.9161} & \underline{1.2949}  \\
\textcolor{gray}{CAT-K}~\cite{zhang2024catk}  & - & \ding{55}    & \textcolor{gray}{0.7846} & \textcolor{gray}{0.4931} & \textcolor{gray}{0.8106} & \textcolor{gray}{0.9177} & \textcolor{gray}{1.3065}\\ \hline
GUMP* ~\cite{hu2024gump}    & GPT-2 Medium~\cite{brown2020language} & \ding{51}   & 0.7431 & 0.4780 & 0.7887 & 0.7404 & 1.6041  \\
SceneStreamer~\cite{peng2025infgen} & Arch. Inspired & \ding{51}  & 0.7731 & 0.4492 & 0.8084 & 0.9127 & 1.4252 \\
\rowcolor{blue!7}
{\methodname{} (Ours)}   & Qwen2.5-0.5B~\cite{yang2025qwen3} & \ding{51}     & \textbf{0.7833} & \underline{0.4905} & \textbf{0.8105} & \textbf{0.9167} & 1.3417  \\
\bottomrule
\end{tabular}
}
 \vspace{-8mm}
\end{table}

\subsubsection{Comparison with SOTAs}
We evaluate \methodname{} on the standard 8s benchmark (Tab.~\ref{tab:comparison}). To align with this short-term setting, we disable the autoregressive agent spawning module during inference, requiring no additional retraining. \methodname{} achieves highly competitive results, outperforming all existing methods except those explicitly optimized with expensive closed-loop reinforcement learning or finetuning. Crucially, it significantly surpasses both LLM-inspired architectures and prior works utilizing pretrained LLMs. Furthermore, among the subset of versatile simulators capable of dynamic agent insertion, \methodname{} establishes a new state-of-the-art by a large margin.
Lastly, while the recent long-term simulator InfGen~\cite{infgen_iccv} is absent from the WOSAC 2025 leaderboard, its reported metrics inherently trail behind SMART~\cite{nips24smart} even on a much smaller validation split. Since \methodname{} strictly outperforms SMART, it deductively establishes superiority over InfGen in this setting as well.

\subsubsection{Ablation over tunable layers}
To investigate whether the performance gains stem from the pretrained linguistic knowledge or merely from increased model capacity, we conduct an ablation study by varying the number of tunable layers within the LLM backbone, while keeping the motion head and adapter layers trainable. As shown in Tab.~\ref{tab:ablation_tunable_layers}, even when the LLM backbone is fully frozen (`None'), our method outperforms the baseline~\cite{nips24smart}, indicating that the pretrained internal representations of the LLM already contain highly transferable structure-awareness. Furthermore, fine-tuning only the last transformer layer (`Last 1') achieves a Composite score of 0.7648, which is highly comparable to the fully fine-tuned result (`All', 0.7658). Interestingly, increasing the tunable scope to the last five layers (`Last 5') does not yield further improvements compared to `Last 1'. This saturation suggests that the pretrained knowledge is robust and requires only minimal alignment to adapt to the traffic simulation.

To explicitly separate the contribution of pretrained knowledge from the underlying transformer architecture, we further compare our model against a randomly initialized Qwen-0.5B (Fig.~\ref{fig:llm_prior_ablation}). We observe that while architectural inductive biases alone account for 63\% of the gain over the baseline, the pretrained statistical prior still contributes the remaining 37\%, validating the distinct importance of LLM pretraining.

\vspace{-5mm}
\begin{table}[htbp]
    \centering
    \begin{minipage}[t]{0.64\textwidth}
        \vspace{0pt}
        \caption{Ablation study on the number of tunable layers within the LLM backbone.
        Only motion head and adapter layers are tunable.
        The experiment is conducted on 2\% of the WOSAC val split.}
        \label{tab:ablation_tunable_layers}
        \resizebox{1\textwidth}{!}{
            \begin{tabular}{lcccc}
            \hline
            \begin{tabular}[l]{@{}l@{}}Tunable \\ Layers \end{tabular}
             & {Composited} $\uparrow$
             & {Kinematic}$\uparrow$
             & {Interactive} $\uparrow$
             & {Map-based}$\uparrow$ \\ \hline
            All                     & 0.7658              & 0.4854             & 0.8095               & 0.8697             \\
            Last 5                  & 0.7647              & 0.4854             & 0.8083               & 0.8682             \\
            Last 1                  & 0.7648              & 0.4876             & 0.8084               & 0.8671             \\
            None                    & 0.7638              & 0.4843             & 0.8074               & 0.8674             \\ \hline
            baseline~\cite{nips24smart}                   & 0.7631              & 0.4829             & 0.8065               & 0.8675             \\ \hline
            \end{tabular}
        }
    \end{minipage}
    \hfill
    \begin{minipage}[t]{0.32\textwidth}
        \vspace{0pt}
        \centering
        \includegraphics[width=\linewidth]{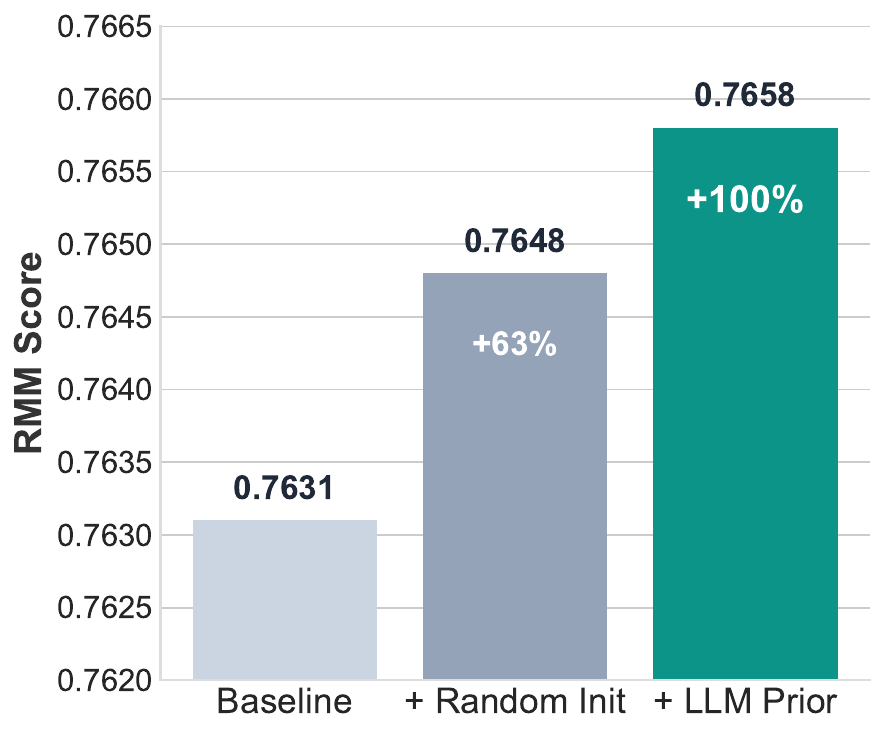}
        \captionof{figure}{Ablation study on LLM's prior. All parameters are tunable.}
        \label{fig:llm_prior_ablation}
    \end{minipage}
\end{table}
\vspace{-15mm}

\subsubsection{Can all LLMs be adapted to traffic simulation?}
To investigate the impact of foundational architectures, we adapt four LLMs under a fixed single-epoch budget to evaluate their {rapid adaptation capability}.
As shown in Tab.~\ref{tab:different_llms}, modern LLMs significantly outperform older architecture like GPT-2, proving that modern LLMs are adaptable.
Notably, {Qwen2.5-0.5B} achieves the best performance, surpassing larger models like Llama-3.2. We hypothesize that Qwen's extensive {multilingual pre-training} fosters a more generalized sequence modeling capability. By treating motion tokens as a ``foreign language'', models exposed to diverse linguistic syntaxes during pre-training can transfer their generalized attention mechanisms more effectively than monolingual dominant models.

\begin{table}[]
    \vspace{-5mm}
    \centering
    \caption{Ablation study on adaptation abilities over different LMs. Evaluated on 2\% of WOSAC \textit{val split}, all the transformer layers on LMs are frozen.}
    \label{tab:different_llms}
    \resizebox{\textwidth}{!}{
\begin{tabular}{lcccclcc}
\hline
\multicolumn{1}{c}{\multirow{2}{*}{LLMs}} & \multicolumn{4}{c}{{WOSAC Metric}} &  & \multicolumn{2}{c}{{Model Card}}                                                                                   \\
\multicolumn{1}{c}{}                      & Composited$\uparrow$   & Kinematic $\uparrow$  & Interactive$\uparrow$   & Map-based$\uparrow$   &  & \begin{tabular}[c]{@{}c@{}}Pretrain \\ Tokens\end{tabular} & \begin{tabular}[c]{@{}c@{}}Supported\\ Languages\end{tabular} \\ \hline
GPT-2-Small~\cite{brown2020language} (2019)                         & 0.5425     & 0.3155     & 0.5590       & 0.6510     &  & -                                                       & -                                                            \\
TinyLlama-1.1B-Chat-v1.0~\cite{zhang2024tinyllama} (2023)           & 0.7397      & 0.4684     & 0.7693       & 0.8566     &  & 3T                                                         & 1                                                            \\
Llama-3.2-1B-Instruct~\cite{grattafiori2024llama} (2024)              & 0.7594      & 0.4817     & 0.8001       & 0.8658     &  & 9T                                                         & 8                                                            \\
Qwen2.5-0.5B-Instruct~\cite{yang2025qwen3} (2025)              & 0.7620      & 0.4835     & 0.8051       & 0.8656     &  & 18T                                                        & 29                                                           \\ \hline
\end{tabular}
    }
    \vspace{-10mm}
\end{table}
\subsubsection{Why LLMs can be adapted to traffic simulation?}
\begin{figure}[htbp]
    \vspace{-4mm}
    \centering
    \begin{minipage}[c]{0.83\textwidth}
        \centering
        \includegraphics[width=0.32\linewidth]{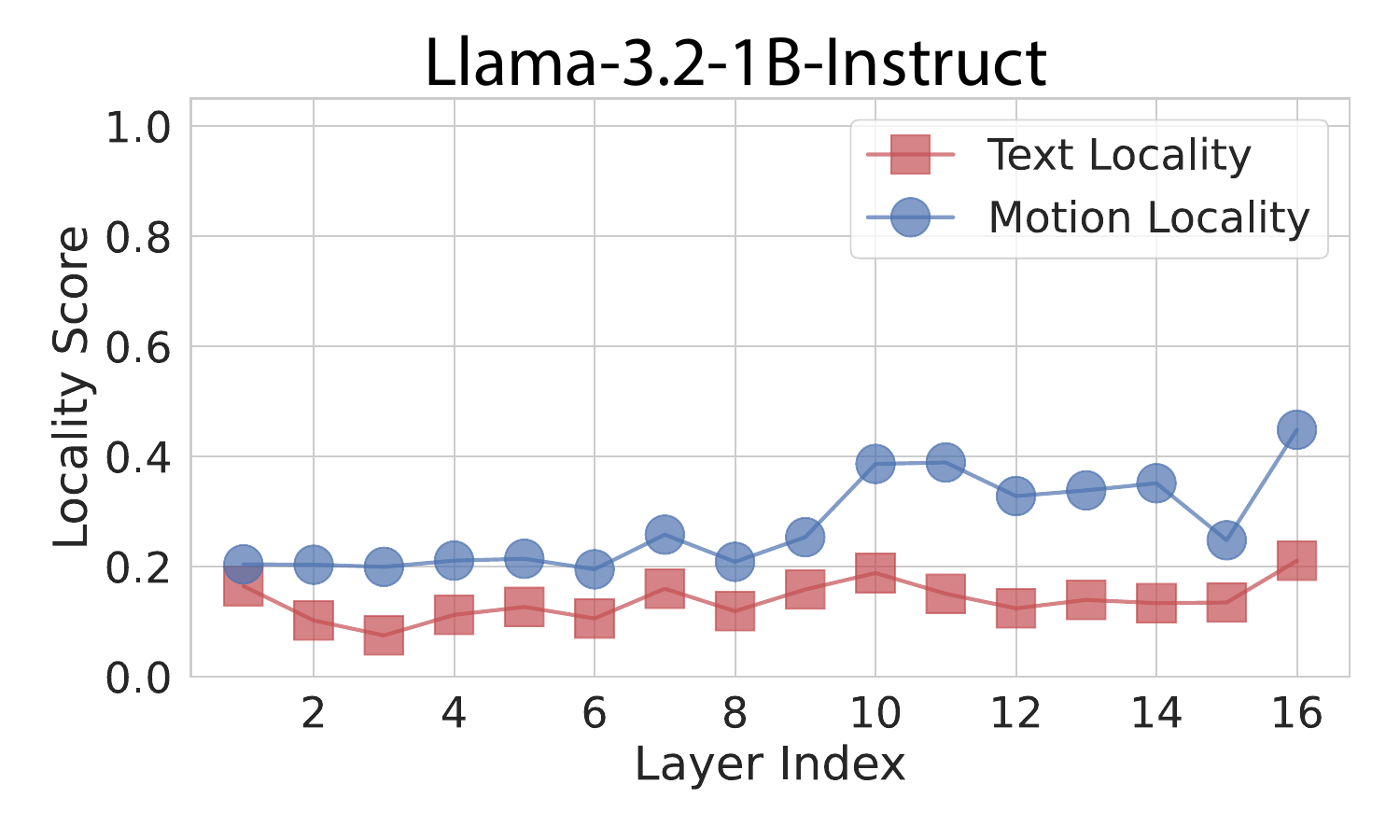}
        \includegraphics[width=0.32\linewidth]{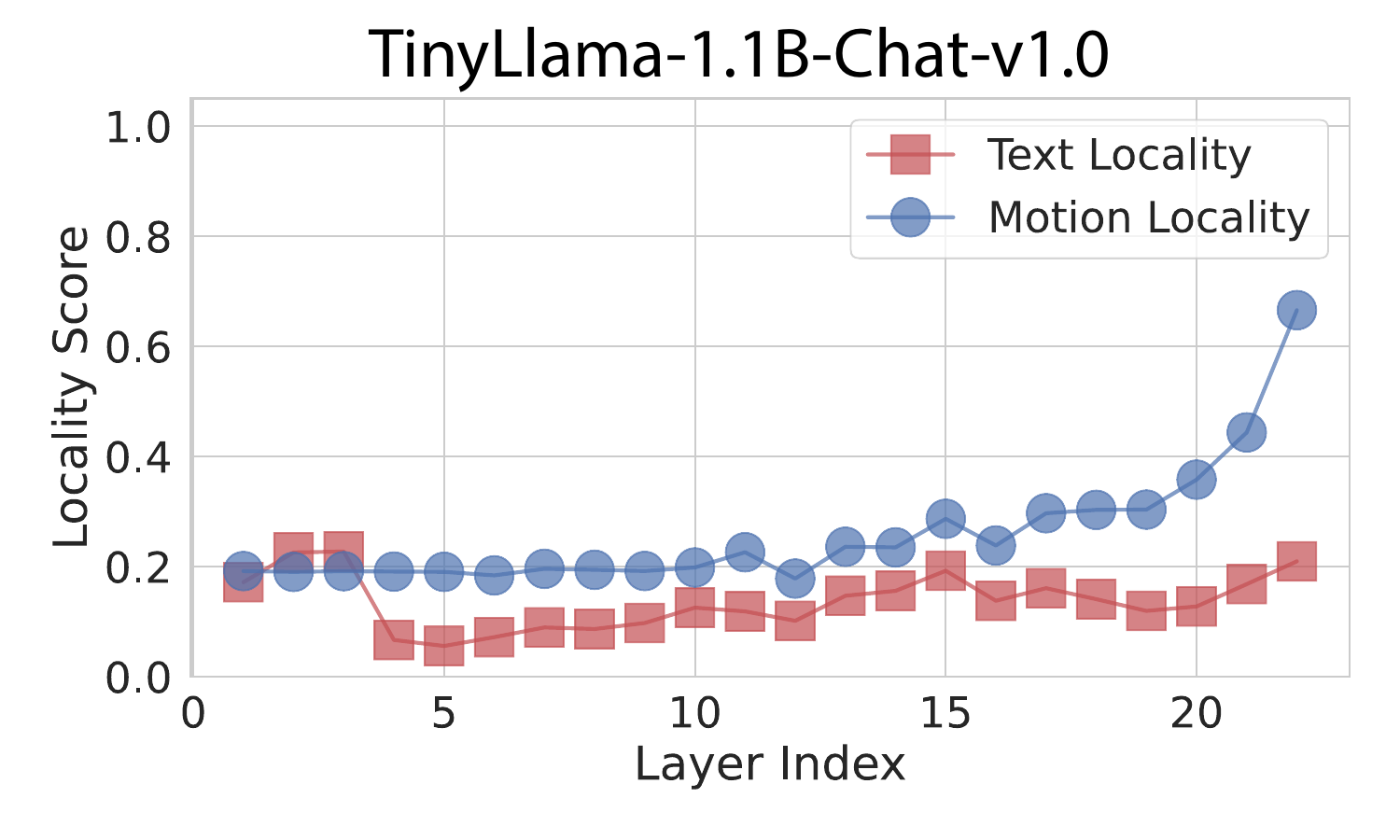}
        \includegraphics[width=0.32\linewidth]{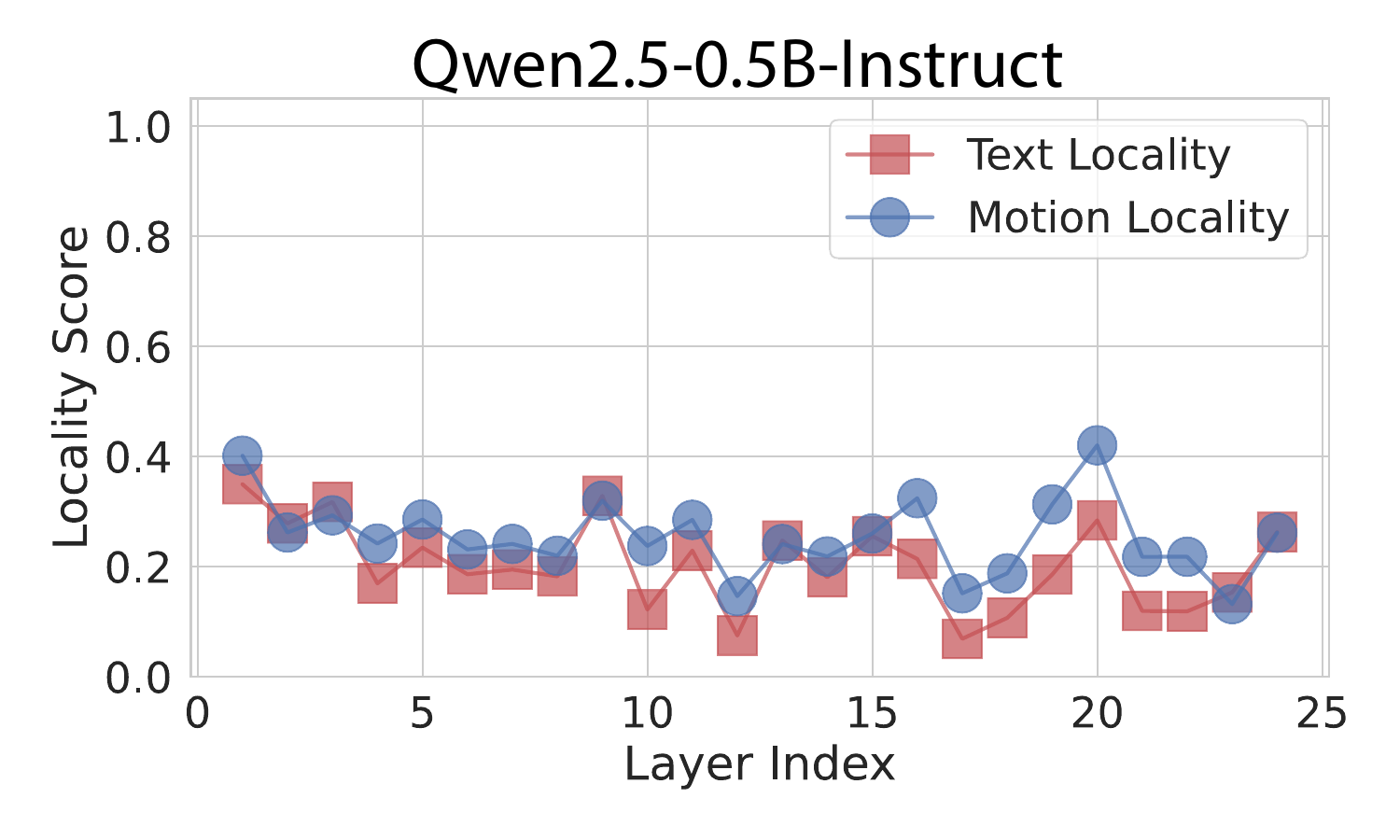}
        \vspace{-3mm}
        \caption{Comparison of the attention mechanism in different LLMs during inference with different sources of data. A higher value means the model is focusing more on adjacent tokens.}
        \label{fig:comparison_attention}
    \end{minipage}
    \hspace{0.01\linewidth}
    \begin{minipage}[t]{0.14\linewidth}
        \centering
        \includegraphics[width=\linewidth]{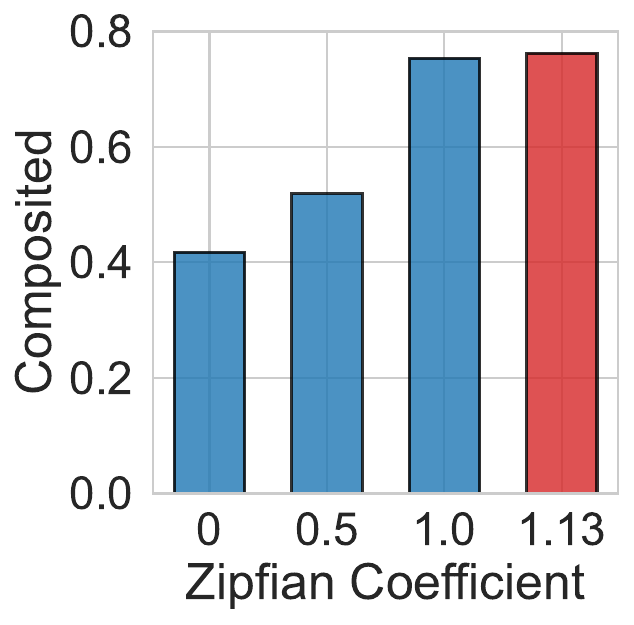}
        \caption{Zipfian corruption.}
        \label{fig:zipfian_corrupt}
    \end{minipage}
    \vspace{-13mm}
\end{figure}

To understand why LLMs can be adapted to traffic simulation, inspired by previous works~\cite{chan2022data}, we manage to explain this in two aspects: 1) the data distribution; 2) the attention mechanism. As shown in Fig.~\ref{fig:zipfian_corrupt}, during the training, we resample the motion tokens and enforce that the distribution follows a zipfian distribution with a different coefficient. We find that when the distribution is not extremely skewed (e.g., a uniform distribution, zipfian coefficient is $0$), the performance will drop significantly, which proves our assumption. In addition, to fully understand why some LLMs are better than others, we visualize the locality of attention during inference under different sources of data across all layers.
As shown in Fig.~\ref{fig:comparison_attention}, the attention locality under text data~\cite{hendrycks2020measuring} and motion data on Qwen2.5-0.5B are similar across different layers, where Llama and TinyLlama are not. These results give us more direct evidence that how LLMs understand language can be metaphorically transferred to how they understand motion tokens.

\subsubsection{Does scaling LLMs help traffic simulation?}
Finally, we want to examine whether scaling up model size benefits traffic simulation, given the intuition that larger models possess richer knowledge. We examined scaling behavior within the same model family (e.g., Qwen2.5 series).
As shown in Tab.~\ref{tab:scaling_freeze}, increasing model size yields no improvement under the frozen setting; in fact, performance drops at 3B. However, a compelling phenomenon emerges when finetuning the last layer: despite the initial performance variance in the frozen setting, the final performance converges across different model sizes (Tab.~\ref{tab:scaling_tune_last_layer} and Tab.~\ref{tab:scaling_different_llms}).

This consistency suggests that the gains from language pretraining primarily stem from general syntactic capabilities, such as modeling sequential dependencies and statistical distributions, rather than deep, specific semantic knowledge. The "grammar" of traffic dynamics is relatively fundamental and can be fully captured by smaller architectures (e.g., 1B-1.5B). Consequently, the advanced reasoning capabilities and specific world knowledge inherent in larger models (e.g., >3B) provide diminishing returns or are even redundant for motion generation. This indicates that traffic simulation relies more on the structural similarity to language than on the complexity of the model's internal knowledge base.
\begin{table}[htbp]
    \vspace{-8mm}
  \centering
  \begin{minipage}[t]{0.28\textwidth}
    \centering
    \caption{Comparison of different model sizes. LLMs are frozen.}
    \vspace{-3mm}
    \label{tab:scaling_freeze}
    \resizebox{\textwidth}{!}{
    \begin{tabular}{lc}
      \hline
      Models & Composited \\ \hline
      Qwen2.5-0.5B   & 0.7620     \\
      Qwen2.5-1.5B   & 0.7620     \\
      Qwen2.5-3B     & 0.7589     \\ \hline
    \end{tabular}
    }
  \end{minipage}
  \hfill
  \begin{minipage}[t]{0.28\textwidth}
    \centering
    \caption{Comparison of different model sizes. Only tune the last layer.}
    \vspace{-3mm}
    \label{tab:scaling_tune_last_layer}
    \resizebox{\textwidth}{!}{
    \begin{tabular}{lc}
      \hline
      Models & Composited \\ \hline
      Qwen2.5-0.5B   & 0.7637     \\
      Qwen2.5-1.5B   & 0.7644     \\
      Qwen2.5-3B     & 0.7641     \\ \hline
    \end{tabular}
    }
  \end{minipage}
  \hfill
  \begin{minipage}[t]{0.30\textwidth}
    \centering
    \caption{Comparison of different model genres. Only tune the last layer.}
    \vspace{-3mm}
    \label{tab:scaling_different_llms}
    \resizebox{\textwidth}{!}{
    \begin{tabular}{lc}
      \hline
      Models & Composited \\ \hline
      Qwen2.5-0.5B   & 0.7637     \\
      TinyLlama-1.1B   & 0.7646     \\
      Llama-3.2-1B   & 0.7650     \\ \hline
    \end{tabular}
    }
    \vspace{-10mm}
  \end{minipage}
\end{table}

\subsection{Long-term Traffic Simulation}
In this section, we want to answer the following questions: 1) \textit{How well does RTE reflect the quality of long-term traffic simulation?} 2) \textit{How does our method perform on long-term traffic simulation?}

\subsubsection{Rollout setup}
Following the practice in~\cite{infgen_iccv}, we set the rollout length to $T=31.1s$; there is no early stopping during rollout. If one ego-vehicle driving segment does not contain any map within a range of $64 \times 64$ throughout the window length $\tau$, we will exclude this from evaluation.

\begin{figure}[h]
     \vspace{-5mm}
    \includegraphics[width=1\textwidth]{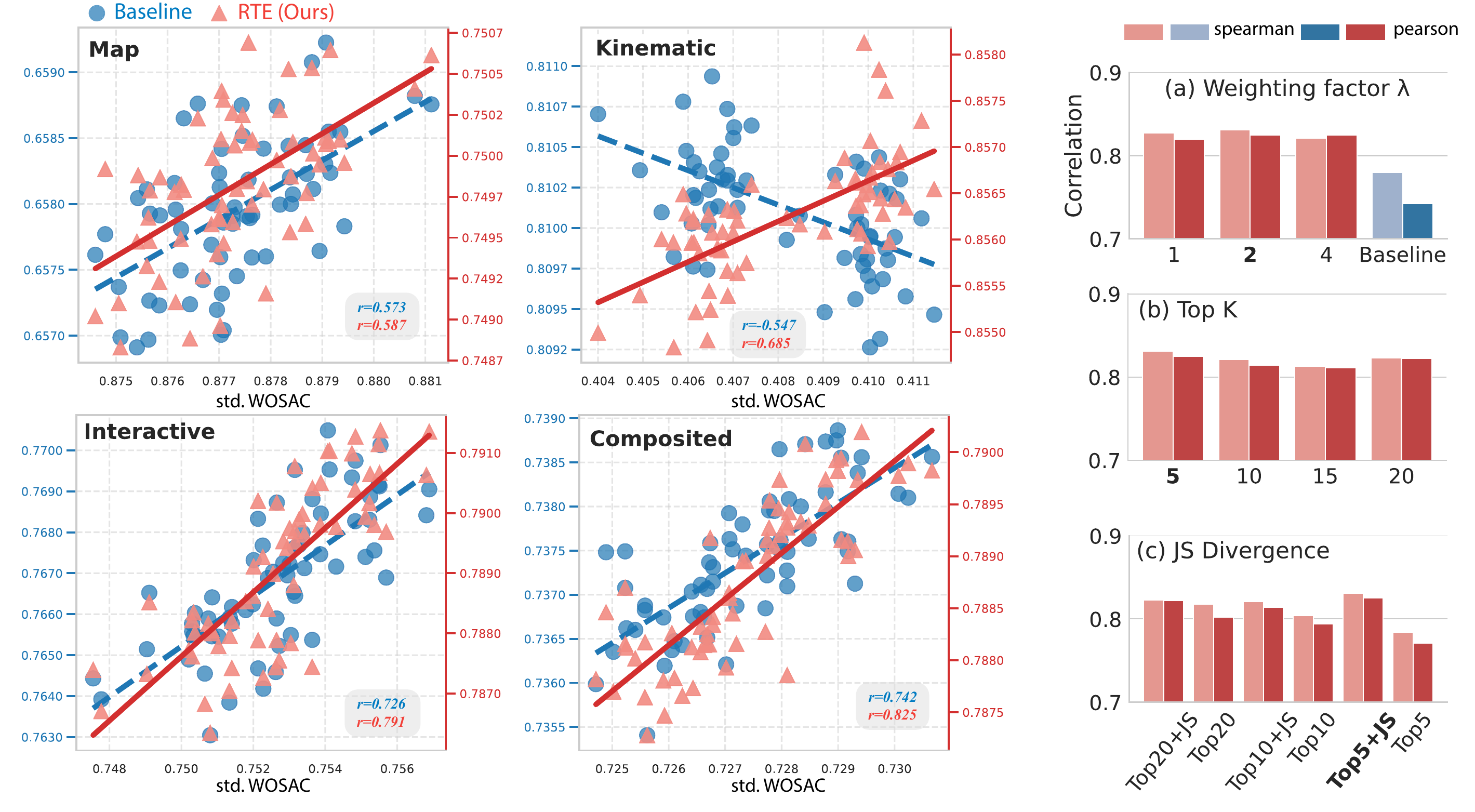}
    \vspace{-5mm}
    \caption{\textbf{Correlations.} Left: Pearson correlations comparison between RTE and log-based metrics against standard metrics. Right: Ablation of RTE's parameters.}
    \label{fig:correlation}
    \vspace{-8mm}
\end{figure}

\subsubsection{How well does RTE reflect the quality of long-term traffic simulation?}
\label{sec:RTE_correlation}

Ideally, metrics designed for long-term simulation should exhibit high correlation with standard short-term WOSAC metrics when evaluated on a single short-term window. To validate this, we generate 63 diverse rollout variants by applying stochastic decoding (top-48 sampling) to two representative base models~\cite{nips24smart,zhang2024catk}. Evaluating these variants on 10\% of the validation set over a standard 8s duration, we compute the Pearson correlation ($r$, with $p \approx 0$) between the standard WOSAC metrics and both our proposed RTE-based metrics and prior log-based metrics.

As shown in Fig.~\ref{fig:correlation} (left), RTE-based metrics (\textcolor{red}{red}) demonstrate significantly higher correlations than log-based metrics (\textcolor{blue}{blue}). This confirms that RTE accurately reflects simulation quality even without strict one-to-one matching. Notably, log-based Kinematic metrics exhibit a \textit{negative} correlation with standard WOSAC metrics. This empirically validates our core motivation (Sec.~\ref{sec:RTE}): blindly forcing specific local scenarios (e.g., low-speed traffic) to match a global kinematic distribution inevitably penalizes realistic behaviors.

Furthermore, Fig.~\ref{fig:correlation} (right) presents an ablation on RTE hyperparameters ($K$ and $\lambda$), demonstrating robust stability across various settings. We also find that applying Jensen-Shannon (JS) divergence to Bernoulli-distributed metrics (e.g., collision and offroad likelihood) further improves overall correlation. This indicates that without a rigid one-to-one assumption, JS divergence effectively smooths distribution discrepancies, yielding a more reliable quality assessment.

\vspace{-3mm}
\subsubsection{Evaluation results}

Tab.~\ref{tab:long-term-comparison} presents the quantitative comparison against InfGen~\cite{infgen_iccv}, CAT-K~\cite{zhang2024catk}, and SMART~\cite{nips24smart}. \methodname{} establishes a new state-of-the-art on the comprehensive RMM-F1 metric (0.7624). Notably, it substantially outperforms the autoregressive baseline, InfGen, in Traffic Flow Realism (0.7846 vs. 0.7637). This validates that our unified architecture effectively manages dynamic population density and maintains global flow consistency over extended horizons.
Regarding Behavior Realism, \methodname{} exhibits outstanding performance in the Kinematic metric, significantly outperforming all baselines.
Rather than overfitting to specific interactive or turning behaviors, \methodname{} achieves a significantly better holistic balance, ensuring robust performance across both behavioral realism and global traffic flow.
Methods lacking long-term optimization often show better behavioral realism because they do not spawn new agents over time. This lower agent density results in fewer collision opportunities, inflating their performance scores.

Furthermore, we report the metrics under $D_{enter}$ and $D_{exit}$ strictly for completeness.
We exclude these metrics from the computation of RMM-F1 score as they can easily exploited by heuristic agent removal. Exhaustive breakdowns under each metrics are deferred to the Supp.

\begin{table}[h]
    \vspace{-4mm}
    \caption{Performance under long-term traffic simulation.}
    \label{tab:long-term-comparison}
    \vspace{-2mm}
    \resizebox{\textwidth}{!}{
\begin{tabular}{lcccccccccccc}
\hline
\multicolumn{1}{c}{\multirow{2}{*}{{Method}}} & \multirow{2}{*}{{RMM-F1}} &  & \multicolumn{5}{c}{{Traffic Flow Realism}} &  & \multicolumn{4}{c}{{Behavior Realism}} \\ \cline{4-8} \cline{10-13}
\multicolumn{1}{c}{}                                 &                              &  & {Overall}     & \#Enter     & \#Exit   & \textcolor{gray}{$D_{enter}$} & \textcolor{gray}{$D_{exit}$} & & {Overall} & Kinematic & Interactive & Map-based \\ \hline
InfGen~\cite{infgen_iccv}       & 0.7290  &  & 0.7637  & 0.7436  & 0.7838  & \textcolor{gray}{0.2734} & \textcolor{gray}{0.1766} & & 0.6973  & 0.7312  & 0.7445  & 0.6172 \\
CAT-K~\cite{zhang2024catk}      & 0.7381  &  & 0.7191  & 0.6862  & 0.7520  & \textcolor{gray}{0.2294} & \textcolor{gray}{0.3167} & & 0.7581  & 0.8582  & 0.8256  & 0.6140 \\
SMART~\cite{nips24smart}        & 0.7383  &  & 0.7175  & 0.6862  & 0.7489  & \textcolor{gray}{0.2291} & \textcolor{gray}{0.3175} & & 0.7602  & 0.8619  & 0.8280  & 0.6150 \\
\rowcolor{blue!7}
\methodname{} (Ours)            & {0.7624}                      &  & 0.7846          & 0.7870    & 0.7821    & \textcolor{gray}{0.2336} & \textcolor{gray}{0.3095} & & 0.7414  & 0.9176  & 0.7663  & 0.6089 \\ \hline
\end{tabular}
    }
    \vspace{-6mm}
\end{table}

\section{Conclusion}
In this work, we propose \methodname{}, a unified framework that seamlessly handles both behavior modeling and traffic flow generation. Our method achieves state-of-the-art performance on both short-term and long-horizon traffic simulation benchmarks. Moreover, we introduce RTE, a novel evaluation protocol that better captures the fidelity of long-horizon traffic simulation.

\vspace{-3mm}
\subsubsection{Limitations and future works }

We rely on heuristics for agent removal for its simplicity, though a separate head or special tokens could easily be integrated later. Future work would introduce properties like initial speeds for spawned agents to improve long-term evaluation and refine the binary collision indicator to account for collision frequency, severity, and pattern~\cite{fengbeyond}.
\newpage
\subsubsection{Acknowledgement}
This work was supported by The University of Hong Kong (HKU) Faculty Interdisciplinary Fund and the HKU Urban Systems Institute (HKU-USI) Fellowship Grant.

\bibliographystyle{splncs04}

\bibliography{arxiv}

\appendix
\setcounter{section}{0}
\setcounter{subsection}{0}
\setcounter{equation}{0}
\renewcommand{\thesection}{\Alph{section}}
\renewcommand{\thesubsection}{\Alph{section}.\arabic{subsection}}
\renewcommand{\theequation}{\Alph{section}.\arabic{equation}}
\renewcommand{\theHsection}{appendix.\Alph{section}}
\renewcommand{\theHsubsection}{appendix.\Alph{section}.\arabic{subsection}}
\renewcommand{\theHequation}{appendix.\Alph{section}.\arabic{equation}}
\section*{Appendix}
\startcontents[sections]
\providecommand{\authcount}[1]{}
\printcontents[sections]{l}{1}{\setcounter{tocdepth}{2}}
\newpage

\section{Table of Notations}
\vspace{-5mm}
\begin{table}[htbp]
\centering
\caption{Table of Notations}
\label{tab:notation}
\renewcommand{\arraystretch}{1.2}
\resizebox{\textwidth}{!}{
\begin{tabular}{cl}
\toprule
\textbf{Symbol} & \textbf{Description} \\
\midrule
\multicolumn{2}{c}{\textit{Indices and Variables}} \\
\midrule
$t, T$ & Current timestep and the extended simulation horizon \\
$i, j$ & Indices for existing agents and newly spawned agents \\
$N_t$ & Number of active agents at timestep $t$ \\
$J$ & Number of newly spawned agents at the current timestep \\
$M_t$ & Number of map tokens \\
\midrule
\multicolumn{2}{c}{\textit{Scene and State Representations}} \\
\midrule
$s_t^i$ & State token (e.g., position, heading, velocity) of the $i$-th agent at time $t$ \\
$\mathcal{S}_t$ & Global traffic scene at time $t$, defined as $\{s_t^1, s_t^2, \dots, s_t^{N_t}\}$ \\
$\tilde{\mathcal{S}}_{t+1}$ & Intermediate scene with updated motions before topological changes \\
$\tilde{s}_{t+1}^i$ & Motion-updated intermediate state of existing agent $i$ \\
$\mathcal{C}$ & Global map context \\
\midrule
\multicolumn{2}{c}{\textit{Token Representations}} \\
\midrule
$a_t^{1:N_t}$ & Agent tokens encoded from the scene context \\
$mo^{1:N_t}, mo_t^{1:N_t}$ & Initial learnable motion tokens and LLM-updated motion tokens \\
$m^{1:M_t}$ & Map tokens extracted from the pretrained map encoder \\
$l^{1:N_t}$ & Layout tokens containing geometrical and topological context \\
$g^i, y^i, c^j$ & Grid token, yaw token, and type token \\
$\text{<SOS>}, \text{<EOS>}$ & Special tokens indicating the start and end of autoregressive generation \\
\midrule
\multicolumn{2}{c}{\textit{Models and Functions}} \\
\midrule
$f_\theta$ & Pretrained large sequence model (autoregressive LLM) \\
$\mathcal{E}$ & Scene encoder \\
$\Gamma(\cdot), \Phi(\cdot)$ & Tokenizers mapping intermediate states to discretized grid and yaw indices \\
$emb_g(\cdot), emb_y(\cdot)$ & Embedding layers for grid and yaw tokens \\
$SinPE(\cdot)$ & Sinusoidal position embedding function \\
$mlp_{(\cdot)}$ & Multilayer perceptrons for downstream projections (motion, type, grid, yaw) \\
\midrule
\multicolumn{2}{c}{\textit{Retrieval-based Traffic Evaluation (RTE)}} \\
\midrule
$\mathcal{E}_\phi$ & Pretrained Scene VAE encoder \\
$\tau$ & Temporal length of the sliced short-term segment \\
$z$ & Latent representation consisting of object-level ($z_{object}$) and map-level ($z_{map}$) \\
$W_2$ & Wasserstein distance used to retrieve semantically similar scenarios \\
\bottomrule
\end{tabular}
}
\vspace{-10mm}
\end{table}
\newpage
\section{Videos}
We provide 30-second rollouts across 27 diverse scenarios for both \methodname{} and the prior state-of-the-art method, InfGen~\cite{infgen_iccv}. The videos have been concatenated into a single visual reel and sped up $6\times$ for better comparative visualization. The actual simulation timestep is displayed in the bottom right corner of the video to provide temporal context.

\section{Additional Related Works}

\subsection{Closed-loop Traffic Simulation}
\subsubsection{Short-term Traffic Simulation}
Traditional traffic simulation~\cite{caesar2021nuplan,dosovitskiy2017carla} relies on hand-crafted rules and physics-based models, which often fail to capture the complex interactions and diverse behaviors of real-world traffic. Recent works have explored data-driven approaches. One early attempt is TrafficSim~\cite{suo2021trafficsim}, which uses a graph-based model to simulate multi-agent interactions.

Subsequent works~\cite{trafficgen,zhong2023guided} either framed the problem as a combination of scene initialization and multi-agent interaction, or pursued controllable multi-agent simulation via diffusion models~\cite{janner2022planning}.

With the launch of the WOSAC benchmark~\cite{montali2023waymo}, the community shifted its attention toward modeling interactions in a more realistic, standardized way.
Early works~\cite{wang2023multiverse,qian20232nd} modified the state-of-the-art motion forecasting model~\cite{shi2024mtr++} for multi-agent simulation tasks. However, because motion forecasting models are optimized under open-loop settings, these approaches often fall short under closed-loop evaluation. Motivated by task formulations in language modeling, Trajeglish~\cite{philion2023trajeglish} proposed modeling multi-agent simulation as a next-token-prediction task and introduced a dedicated tokenization strategy. Expanding on this, SMART~\cite{nips24smart} proposed an autoregressive architecture that proved highly scalable for traffic simulation. To mitigate distributional shifts between open-loop training and closed-loop testing, CAT-K~\cite{zhang2024catk} proposed a closed-loop fine-tuning strategy to bypass extremely time-consuming rollouts and reward shaping. Following the release of GPU-accelerated simulators~\cite{waymax,kazemkhani2024gpudrive}, several works~\cite{guo2025decompgail,pei2025advancing,ahmadi2025rlftsim,chang2025spacer,cusumano2025robust} applied reinforcement learning to further boost performance.

Other streams of research have investigated architectural designs. BehaviorGPT~\cite{zhou2024behaviorgpt} formulated the task as `next-patch-prediction' to build a parameter-efficient model. KiGRAS~\cite{zhao2024kigras} modeled the task in physical action space instead of state space, reducing task complexity and ensuring physical feasibility. UniMM~\cite{lin2025revisit} revisited past design choices to propose a unified framework, while LLM2AD~\cite{wang2025llm} investigated whether design concepts from language models could be conceptually applied to simulation. comBOT~\cite{combot2025ensemble} proposed an ensemble method combining multiple approaches, and TrajTok~\cite{zhangtrajtok} improved tokenization strategies. Additionally, works like~\cite{tan2024promptable,chang2025langtraj} have leveraged language as a conditioning mechanism for simulation.

The closest work to ours in this domain is GUMP~\cite{hu2024gump}, which leverages a multimodal autoregressive model as the core generative engine for scenario generation and traffic simulation. In terms of short-term simulation, our work structurally differs because: 1) we do not rely on any language priors or instructions, and 2) we formulate the input as a homogeneously structured token sequence rather than mixing multiple modalities, yielding superior performance gains.

\subsubsection{Long-term Traffic Simulation}
Building upon SceneDiffuser~\cite{jiang2024scenediffuser}, SceneDiffuser++~\cite{Tan_2025_CVPR} extended simulation into long-term horizons by formulating the task as jointly modeling scene generation and multi-agent interactions via diffusion. Similarly, the recent work InfGen~\cite{infgen_iccv} follows this joint formulation but implements it under temporal autoregressive modeling, extending upon SMART~\cite{nips24smart}.

Our work is fundamentally different from InfGen as we operate under {spatial-temporal} autoregressive modeling via structured tokenization. Specifically, we perform dynamic agent insertion using structured spawn and layout tokens to enable agent-level autoregressive insertion. This approach is highly flexible and captures complex, asynchronous interactions between agents over time. In contrast, InfGen performs agent insertion by directly spawning entire scene topologies at once, which frequently generates unrealistic traffic flows (e.g., too many agents abruptly entering the scene simultaneously).

\subsection{Traffic Scenario Generation for Simulation}
This line of work focuses on generating initial scenes for traffic simulators. One stream targets the generation of initial agent states~\cite{lu2024scenecontrol,tan_2021_scenegen,peng2025infgen,trafficgen}, map layouts~\cite{mi2021hdmapgen}, or both~\cite{chitta2024sledge,sun2024drivescenegen,rowe2025scenario}. Furthermore, works like LCTGen~\cite{tan2023language,hu2024gump} use language as a condition for initial agent state generation, while RealGen~\cite{ding2023realgen} uses query tokens.

Among these, works like~\cite{peng2025infgen,sun2024drivescenegen, trafficgen, hu2024gump} attempt to explicitly factorize simulation into scene generation and subsequent multi-agent rollouts using a single model, theoretically allowing the simulation to continue indefinitely. However, this factorization is inherently limited because the agent insertion timing is rigid, failing to reflect the naturally evolving, continuous nature of traffic.

Different from these approaches, \methodname{} performs dynamic agent generation during the continuous rollout via autoregressive modeling. This dynamically captures the evolving nature of traffic, yielding significantly better long-term simulation fidelity.

The closest work to ours in this domain is SceneStreamer~\cite{peng2025infgen}, which first generates an initial scene autoregressively and subsequently decodes agent motions in parallel. In contrast, to accurately model traffic evolution, \methodname{} strictly enforces a structured sequence: we first perform motion updates and then conditionally generate the new scene topology. This structured design offers several immediate benefits:
\begin{enumerate}
    \item \textbf{Stability:} Unlike SceneStreamer, where complex token mixtures easily lead to model collapse over long sequences, our method infers agent motions strictly based on previously generated scene tokens, ensuring long-horizon stability.
    \item \textbf{Scalability:} \methodname{} operates under standard causal attention, where sequence relationships are easily captured and newly introduced agents are seamlessly incorporated. SceneStreamer, conversely, relies on complex mixed-attention mechanisms dependent on various token types.
    \item \textbf{Training Efficiency:} Because all tokens are heavily structured, we effectively exploit pre-trained LLM priors. Unlike SceneStreamer, which requires multi-stage training (pretraining and fine-tuning), \methodname{} is optimized in a single stage. Notably, our model, trained for just a few epochs, outperforms SceneStreamer (trained for a total of 35 epochs) on short-term simulation benchmarks.
\end{enumerate}

\subsection{Language Models for Non-language Tasks}
Based on the source of supervisory signals, this line of work can be categorized into two streams: 1) using language as a condition for non-language tasks, and 2) applying language models directly without any language supervision.

The first category uses language as a bridge or formulates a multi-task optimization problem (e.g., vision-language-action tasks in robotics~\cite{kim2024openvla,bjorck2025gr00t,zitkovich2023rt} and autonomous driving~\cite{xu2024drivegpt4,hwang2024emma,renz2025simlingo}). Works like~\cite{dinh2022lift} use natural language to translate non-language tasks into language formats to perform in-context learning.~\cite{shen2023cross} discusses how to transfer pre-trained knowledge between different modalities via a two-stage strategy.

The second category aims to directly apply language models to non-language tasks without any language supervision. For example,~\cite{lu2022frozen} froze an entire language model and only fine-tuned its MLP head and projection layers, demonstrating strong performance on diverse tasks like image classification and numerical computation.~\cite{mirchandani2023large} further investigated how LLMs can complete complex sequential tasks—ranging from probabilistic context-free grammars (PCFG) to spatial patterns in the Abstraction and Reasoning Corpus (ARC). Other works~\cite{pangfrozen} revealed that pre-trained language model layers can be applied directly to vision tasks without vision-specific pretraining, and~\cite{tsimpoukelli2021multimodal} achieved few-shot learning using frozen language models. Building upon this, LatentLens~\cite{krojer2026latentlens} proposed projecting vision tokens into the language space to leverage LLM interpretability. Theoretically,~\cite{chan2022data} explained that the emergent generalization ability of large language models stems from the implicit learning of underlying data structures, which inherently applies to non-language tasks.

Our work belongs to this second category and is the first to rigorously investigate the effectiveness of language models in close-loop traffic simulation.

\section{Training and Inference Details}
\subsection{Tokenization}
\subsubsection{Map Tokenizer}
We adapt the map tokenizer from~\cite{zhang2024catk,nips24smart}, which samples the map into a set of road vectors with a fixed length. Each vector contains fundamental properties such as location, orientation, and road type.

\subsubsection{Motion Tokenizer}
We follow the tokenization scheme established by previous works~\cite{zhang2024catk,nips24smart,guo2025decompgail,pei2025advancing,ahmadi2025rlftsim,philion2023trajeglish}. The continuous trajectories from the dataset are segmented into fixed 0.5-second intervals. K-means clustering is then applied to extract a standardized motion vocabulary. During training, continuous trajectories are discretized and assigned the index of their closest motion vocabulary cluster.

\subsubsection{Grid Tokenizer}
Motivated by~\cite{infgen_iccv}, we construct a spatial grid with a 75-meter radius under an ego-centric coordinate system to serve as our grid vocabulary. The grid resolution is set to 3 meters, and the spatial index of a spawned agent is assigned based on minimum L2 distance.

\subsubsection{Yaw Tokenizer}
We construct the yaw tokenizer under an agent-centric coordinate system. The $360^{\circ}$ space is equally divided into $3^{\circ}$ increments, resulting in a vocabulary size of 120. Similar to the grid tokenizer, the index is assigned based on minimum angular distance.

\subsubsection{Type Tokenizer}
We consider three types of agents: cars, pedestrians, and cyclists, which are assigned indices 0, 1, and 2, respectively. Additionally, we train the $\texttt{mlp}_{type}$ head to predict the \texttt{<EOS>} token (assigned index 3) to dynamically halt the autoregressive spawning process.

\subsection{Hardware and Training Details}
The models reported in the main tables were trained for 7 epochs on 8 RTX 5090 GPUs, utilizing Qwen2.5-0.5B-Instruct as the autoregressive model. The batch size was set to 2 per GPU. The learning rate was set to 3e-4 with a cosine decay schedule and a 0.01 warmup ratio. The training time for each epoch was approximately 10 hours due to the extended token lengths required for full sequence generation. Notably, this is highly efficient compared to SceneStreamer, which required 12 hours per epoch on 8 RTX A6000s during fine-tuning alone. If agent generation is disabled during training (for short-term only applications), the training time drops to roughly 3 hours per epoch.

\subsection{Inference Details}
\subsubsection{Short-term Benchmark}
To ensure strict fairness, we do not incorporate any heuristic strategies during inference for the short-term benchmark. We use top-48 sampling for motion generation.

\subsubsection{Long-term Benchmark}
\label{sec.inference_detail_long}
As \methodname{} does not utilize a dedicated head to predict the continuous offset distance between the discretized grid position and the actual coordinates~\cite{trafficgen,peng2025infgen,infgen_iccv}, we apply a simple heuristic strategy during inference. Specifically, after decoding the grid token into a spatial coordinate, we check if the closest lane centerline point is within 3 meters. If so, we snap the position to the centerline; otherwise, we retain the raw grid position. This lightweight strategy eliminates the need for an additional offset prediction head, further accelerating training. Furthermore, if a newly spawned agent overlaps with existing agents in the scene, the model directly samples the \texttt{<EOS>} token to gracefully halt generation.

For agent deletion, we apply a boundary-based heuristic: any agent exiting the observable map boundary is systematically removed from the scene context, likewise SMART and CAT-K. Additionally, if an agent generates an extreme or physically unrealistic motion token (e.g., an impossibly high acceleration), that agent is also removed to maintain overall traffic stability.

We use top-5 sampling for motion generation and top-1 sampling for grid and yaw generation. For the agent type head, we use top-2 sampling with a variable temperature parameter to control traffic density. Adjusting the sampling temperature for agent generation allows flexible control over scene density: higher temperatures yield denser, more aggressive traffic generation, while lower temperatures produce lighter traffic flows.

\section{Metrics}
\subsection{Standard WOSAC Metric}
All metrics under the short-term setting are calculated by standard WOSAC metrics defined in~\cite{montali2023waymo}. We utilized the open-source codebase from CAT-K~\cite{zhang2024catk} to compute these metrics.

\subsection{Long-term Metrics}

For traffic flow realism, the calculation aligns with~\cite{infgen_iccv} and is implemented using their codebase, including the calculation for the log-level distribution of the entire validation set. However, instead of using NLL-likelihood, we use JS divergence for the \#Enter and \#Exit. We found that when these metrics were evaluated using NLL-likelihood, scenarios with zero entering agents incorrectly received the highest scores—a mathematical artifact of NLL on sparse distributions. Full proof and analysis can be found on Sec.~\ref{sec.supp-placement-math}.

All metrics evaluating behavior realism are calculated against the aggregated log-likelihood of the retrieved scenarios. For collision and offroad likelihoods, we further average the NLL-likelihood with the Jensen-Shannon (JS) divergence between the rollout distribution and the retrieved scenario distribution, as we empirically found this improves correlation with true realism during ablation studies. For kinematic related metric, we calculate the metric under JS divergence instead of NLL due the same mathematical reason.

We calculate these metrics over sliding evaluation windows with a shift step of 5 frames (0.5 seconds at 10Hz). The final metric is computed by first averaging across all windows for a given rollout, and then averaging across all distinct rollout scenarios.

\section{Additional Results on Long-term Traffic Simulation}

\subsection{Is the Retrieval Model Faithful Enough?}
\begin{figure}[htbp]
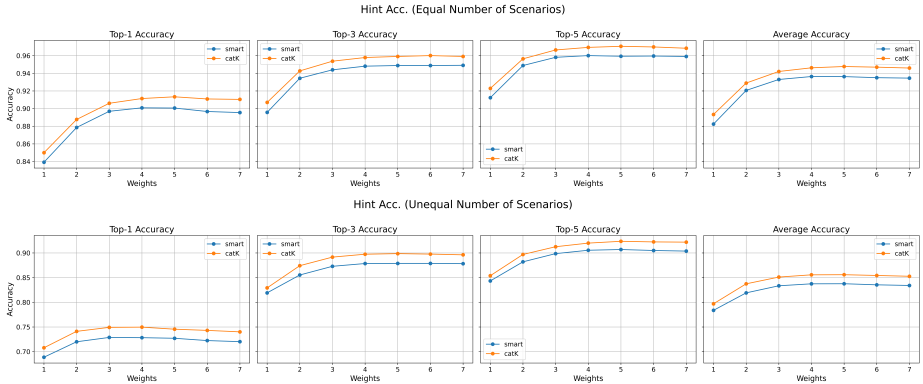

    \centering
    \includegraphics[width=1\textwidth]{figs/supp/retrieval_accuracy_equal.png} \\
    \includegraphics[width=1\textwidth]{figs/supp/retrieval_accuracy_unequal.png}
    \caption{Evaluation of the retrieval model's faithfulness. Top: Evaluated with an equal number of scenarios. Bottom: Evaluated with an unequal number of scenarios.}
    \label{fig:retrieval_acc}
\end{figure}

This section validates whether the retrieval model $\mathcal{E}_{\phi}$ is faithful enough to reliably retrieve the most semantically similar real-world scenarios, which is the foundational premise of our long-term metrics. If the retrieval model fails to find accurate matches, the log-likelihood of the retrieved scenarios cannot reflect the realism of the rollout, rendering the Retrieval-based Traffic Evaluation (RTE) framework invalid.

To verify this, we conducted a rigorous sanity check evaluating retrieval accuracy across different rollout policies. Specifically, we used two different simulation policies to rollout 10\% of the validation set (under the short-term setting), encoded all rollout segments into the latent space, and built a retrieval database from the corresponding ground-truth log trajectories. The core premise is: given an encoded segment $\mathbf{z}^k$ generated by a policy, can the retrieval model find the exact corresponding ground-truth segment $\hat{\mathbf{z}}^k$ from the database?

For each rollout segment, we retrieved the top-K similar segments from the database and computed accuracy based on whether at least one retrieved segment belonged to the same original driving scenario. We conducted this experiment under two strict settings: 1) The total number of rollout scenarios exactly matched the retrieval database (less distraction); and 2) The number of rollout scenarios was significantly smaller than the retrieval database (high distraction).

The results, shown in Fig.~\ref{fig:retrieval_acc}, demonstrate that under both settings, our retrieval model achieves highly robust accuracy (e.g., averaging ~0.90 in the first setting and ~0.85 in the second) across varying distance weights ($\lambda$). This conclusively proves the model is faithful and precise in retrieving structurally analogous reference scenarios.

\subsection{Why use Harmonic Mean?}
\begin{wrapfigure}{r}{0.3\textwidth}
  \centering
  \includegraphics[width=\linewidth]{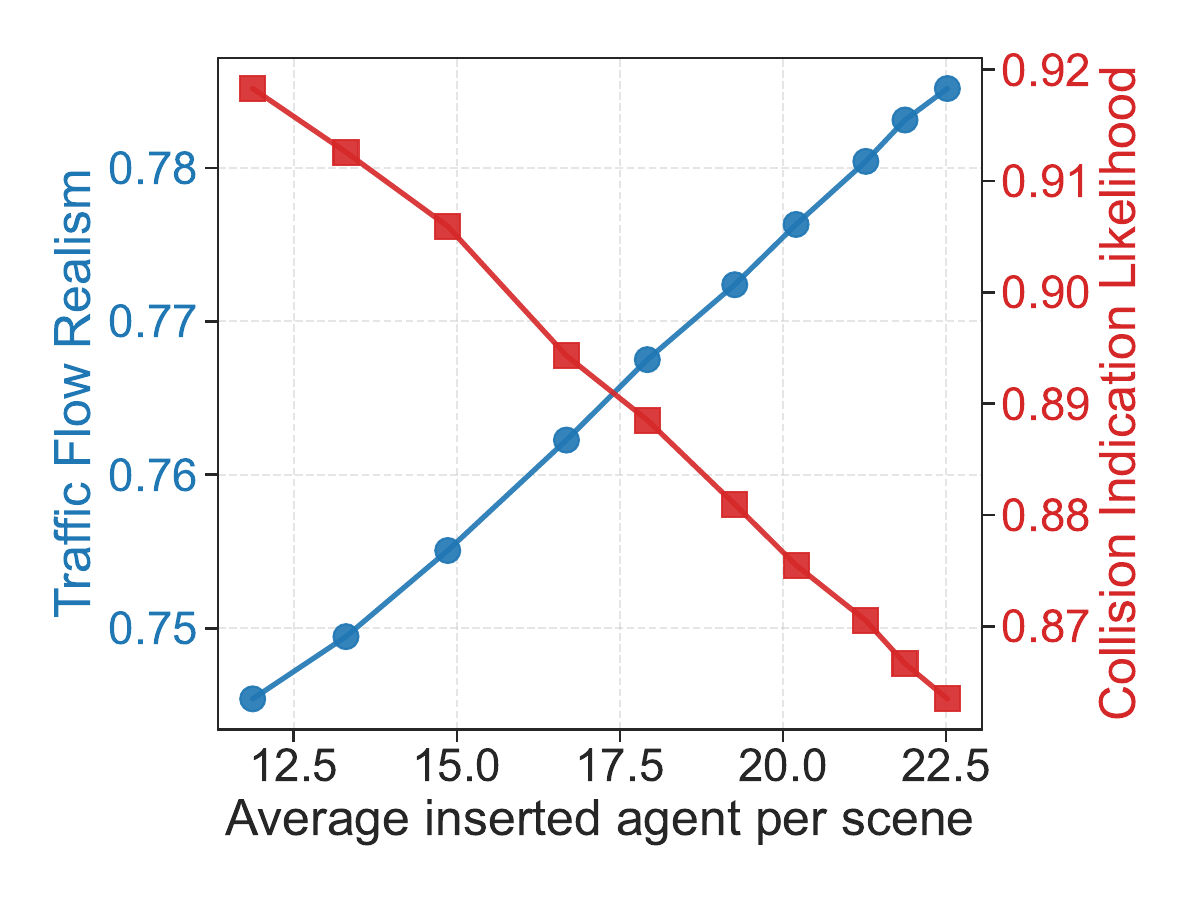}
  \caption{Traffic density vs. Collision indication.}
  \label{fig:density}
\end{wrapfigure}

We plot the relationship between traffic density (flow) and critical behavior metrics (e.g., collision likelihood) in Fig.~\ref{fig:density}. As expected, as traffic density increases, the collision likelihood also decreases, because denser traffic naturally induces more multi-agent interactions and potential conflicts.

However, if we were to use a simple arithmetic mean to combine Behavior Realism and Traffic Flow Realism, the overall evaluation metric would be highly susceptible to exploitation. A model could easily ``game'' the score by aggressively optimizing for one dimension while completely neglecting the other (e.g., achieving high traffic density realism but disastrous collision rates, or vice versa).

Therefore, we employ the harmonic mean to rigorously synthesize these two dimensions. The harmonic mean strictly penalizes models that underperform in either dimension, ensuring that high scores are exclusively awarded to models that achieve a holistic balance between micro-level behavioral realism and macro-level population dynamics, mirroring the foundational logic of the standard F1-score.

\subsection{Why is RTE Excluded from Traffic Flow Metrics?}
\begin{figure}[htbp]
    \centering
    \includegraphics[width=\textwidth]{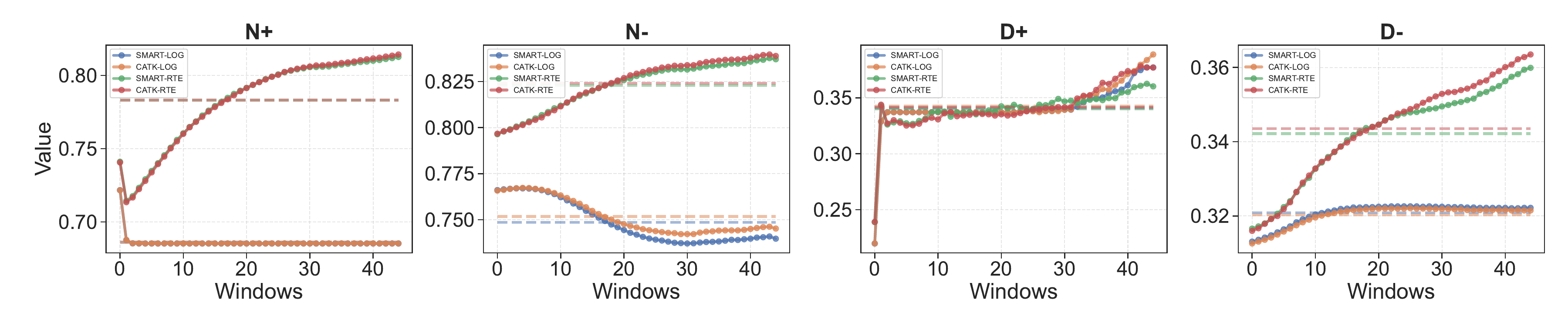}
    \caption{Comparison between RTE placement metrics and log-based placement metrics across rollout windows for short-term specific baselines~\cite{nips24smart,zhang2024catk}. Best viewed zoomed in.}
    \label{fig:placement}
\end{figure}

In Fig.~\ref{fig:placement}, we plot the per-window performance of two short-term specific baselines using both RTE-based and standard log-based calculations for number of entries (N+), exits (N-), entry distances (D+), and exit distances (D-).

Under the RTE-based evaluation, the scores for insertion (N+) and deletion (N-) counts inappropriately increase as the simulation progresses. This is fundamentally incorrect because these baseline models lack the capacity to model traffic evolution entirely. This anomaly can be attributed to the inherent boundaries of the retrieval model's objective. The retrieval model seeks the most semantically similar real-world scenario; as the non-generative baselines slowly empty out (due to agents naturally leaving the scene without being replaced), the retrieval model is forced to pull increasingly sparse scenarios from the database. Consequently, macro-flow properties like the number of inserted agents can appear to ``improve'' over time, even though the underlying simulation is actually degrading.
This proves that log-based metrics are better suited for tracking global density limits, while RTE should be strictly reserved for evaluating nuanced, context-aware interaction fidelity.

\subsection{The Flaw in $D_{enter}$ and $D_{exit}$}
As illustrated in Fig.~\ref{fig:placement}, the entry distance ($D+$) and exit distance ($D-$) scores for non-generative baselines constantly ``improve'' over time under both log-based and RTE-based calculation pipelines. This trend is highly problematic and counter-intuitive.

Conceptually, because models like SMART and CAT-K lack the ability to dynamically spawn new agents or seamlessly delete existing ones mid-scene, their insertion and exit events are artificially constrained to the outermost boundary of the observable range (e.g., the 75-meter threshold). This heuristic limitation manifests identically for both entering and exiting agents, exploiting the metric:
\begin{itemize}
  \item \textbf{For $D_{enter}$ ($D+$):} Because the models cannot realistically spawn vehicles in occluded mid-scene areas (e.g., a driveway), any vehicle entering the simulation is simply an agent crossing the 75-meter threshold from the outside. Thus, $100\%$ of their placement distances are artificially mapped to the outermost spatial bin.
  \item \textbf{For $D_{exit}$ ($D-$):} Similarly, simulated agents never dynamically disappear mid-route. They only exit the simulation naturally by driving beyond the 75-meter boundary, forcing $100\%$ of the exit distances into the maximum bin.
\end{itemize}

As the simulation progresses across sequential time windows, agents naturally leave the map. Without autoregressive mid-scene generation to replenish the traffic, the environment gradually empties. Consequently, the only entry and exit events the model outputs are those strictly tied to the absolute map boundary.

As mathematically proven in Sec.~\ref{sec:metric_exploitation}, this outermost boundary bin naturally contains the highest ground-truth probability mass (e.g., $>33\%$). Because the pointwise log-likelihood metric heavily rewards hitting this specific bin, non-generative models artificially inflate their scores through mode collapse. Simply put, the metric deceptively reports improving performance as the simulated traffic unrealistically disappears. This empirical anomaly validates our theoretical proof: $D_{enter}$ and $D_{exit}$ reward boundary mode collapse rather than evaluating true generative diversity, thoroughly justifying their exclusion from the RMM-F1 protocol.

\subsection{Detailed Metrics under Behavior Realism}

Regarding Behavior Realism, we provide a granular breakdown across the full validation set in Tab.~\ref{tab:detailed_behavior_realism} to elucidate the underlying performance drivers. \methodname{} consistently dominates the Kinematic metrics, underscoring its superior capability in generating physically realistic and smooth vehicle trajectories. This is heavily evidenced by its state-of-the-art performance across all fundamental motion indicators.
While CAT-K and SMART secure higher overall Interactive scores—driven primarily by their collision avoidance (approx. 0.96) and Time-to-Collision (TTC) metrics, \methodname{} remains highly competitive (e.g., TTC of 0.9440) and still significantly outperforms InfGen. It is crucial to note that methods lacking long-term optimization often show better interactive and safety realism because they do not continuously spawn new agents over extended horizons. This artificially lower agent density results in fewer complex multi-agent negotiations and collision opportunities, inevitably inflating their performance scores in these specific metrics.
Furthermore, concerning map-based constraints (e.g., offroad compliance and distance to road edge), all methods exhibit comparable adherence to road boundaries.
\begin{table}[htbp]
    \caption{Detailed behavior realism metrics comparison across the \textit{full validation set}}.
    \vspace{-3mm}
    \label{tab:detailed_behavior_realism}
    \resizebox{1\linewidth}{!}{
    \begin{tabular}{lccccccccc}
    \toprule
    Method & \begin{tabular}[c]{@{}c@{}}angular\\ acc\end{tabular} & \begin{tabular}[c]{@{}c@{}}angular\\ speed\end{tabular} & \begin{tabular}[c]{@{}c@{}}linear\\ acc\end{tabular} & \begin{tabular}[c]{@{}c@{}}linear\\ speed\end{tabular} & collision & \begin{tabular}[c]{@{}c@{}}dist.\\ to obj.\end{tabular} & TTC. & \begin{tabular}[c]{@{}c@{}}dist. to\\ road edge\end{tabular} & offroad  \\ \midrule
    InfGen & 0.8379 & 0.8949 & 0.5716 & 0.6206 & 0.8461 & 0.3128 & 0.9223 & 0.2312 & 0.7717 \\
    CAT-K  & 0.9405 & 0.8892 & 0.9131 & 0.6898 & 0.9602 & 0.3559 & 0.9589 & 0.2267 & 0.7689 \\
    SMART  & 0.9443 & 0.8940 & 0.9153 & 0.6940 & 0.9628 & 0.3606 & 0.9583 & 0.2285 & 0.7696 \\
    \rowcolor{blue!7}
    \methodname{}   & 0.9588 & 0.9426 & 0.9319 & 0.8370 & 0.8786 & 0.3075 & 0.9440 & 0.2266 & 0.7618 \\
    \bottomrule
    \end{tabular}
    }
    \vspace{-10mm}
\end{table}

\subsection{Performance Stability Across Evaluation Windows}
\begin{figure}[htbp]
    \vspace{-5mm}
    \centering
    \includegraphics[width=\textwidth]{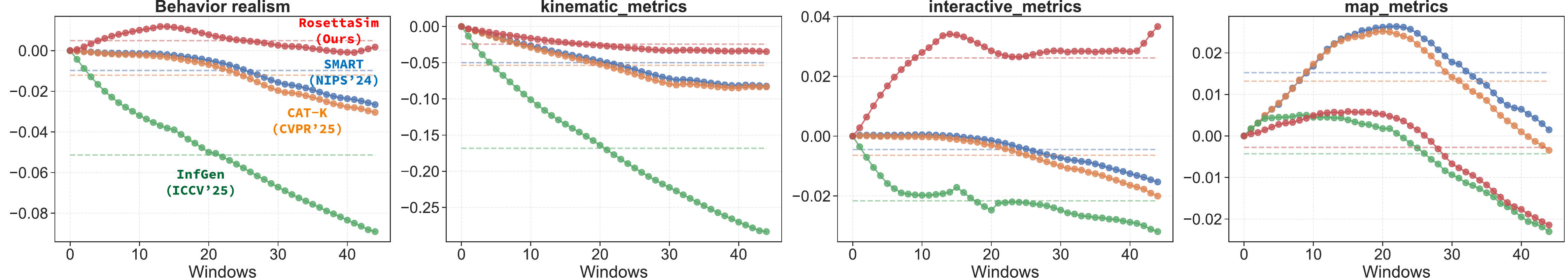}
    \caption{Relative performance changes regarding Behavior Realism across progressive evaluation windows. Best viewed zoomed in. \methodname{} demonstrates superior temporal stability compared to baselines.}
    \label{fig:long-term comparison}
    \vspace{-10mm}
\end{figure}

To evaluate performance stability over time, we report the per-window relative performance divergence compared to the first evaluation window, as shown in Fig.~\ref{fig:long-term comparison}. Due to actively increasing traffic density via autoregressive spawning, our method slightly sacrifices absolute pointwise behavior scores, but critically, it maintains a much more stable performance trajectory throughout the extended horizon.

Our method struggles slightly on map-based metrics primarily because we omitted a dedicated offset-prediction head (relying instead on simple heuristic snapping), newly generated agents may be overlapped with road boundaries. However, compared to InfGen, whose performance gradually decays over the windows, our method's stability gradually improves, proving that \methodname{} successfully captures the evolving nature of traffic. Non-generative methods (SMART, CAT-K) cannot maintain stable performance over long horizons, drastically dropping as their static scenes degrade, further validating that our RTE-based metrics are fair and highly discriminative for long-term modeling.

\subsection{Ablation on heuristic}
As stated on Sec.~\ref{sec.inference_detail_long} we have three heuristic adaptations: (1) centerline snap (2) overlap rejection, and (3) boundary removal. Disabling overlap rejection makes RMM-F1 drop from 0.7346 to 0.7339. It remains the same after further disabling centerline snap. Disable heuristic removal is not feasible as it leads to infinite agent spawning and scene collapse.

\subsection{Robustness against VAE Manifold Bias}
To address the concern that the pretrained Scene VAE introduces an inductive bias (measuring "closeness to the VAE prior"), we clarify that the VAE is strictly used as a retrieval engine to select reference scenarios, not for metric computation. To eliminate potential manifold bias, we replaced the VAE latent space with an orthogonal, bias-free metric: pure $L_2$ distance matching on raw trajectories. As shown in Tab~\ref{tab.supp:ablate_vae} (Rows 1 and 2), despite removing the learned prior, the relative performance ranking in overall behavior realism remains entirely consistent, proving RTE reflects true interactive realism.Furthermore, we conducted retrieval on less similar scenes to simulate a degraded retrieval process (Row 3). In this distorted setting, the rankings break down, and InfGen falsely becomes the top model. This highlights that accurate semantic retrieval via the VAE is indispensable for a fair evaluation of long-horizon behavioral realism.
\begin{table}[hp]
  \centering
  \caption{Ablation study on retrieval methods.}
  \label{tab.supp:ablate_vae}
  \resizebox{1\linewidth}{!}{
  \begin{tabular}{cccc}
  \hline
  Retrieval Methods  & Rank1  & Rank2 & Rank3 \\ \hline
  VAE Latent Space  & SMART (0.7072)   & infgen (0.7021) & RosettaSim (0.6901)        \\
   Ego-MSE Distance & SMART (0.6935) & infgen (0.6900) & RosettaSim (0.6846) \\
   VAE Latent Space (mask out top80\%)     & infgen (0.7052) & SMART (0.7015) & RosettaSim (0.6931)  \\  \hline
  \end{tabular}
  }
\end{table}

\section{Qualitative Analysis}
\subsection{Visualization of Retrieved Scenarios}
\begin{figure}[htbp]
    \centering
    \includegraphics[width=0.9\textwidth]{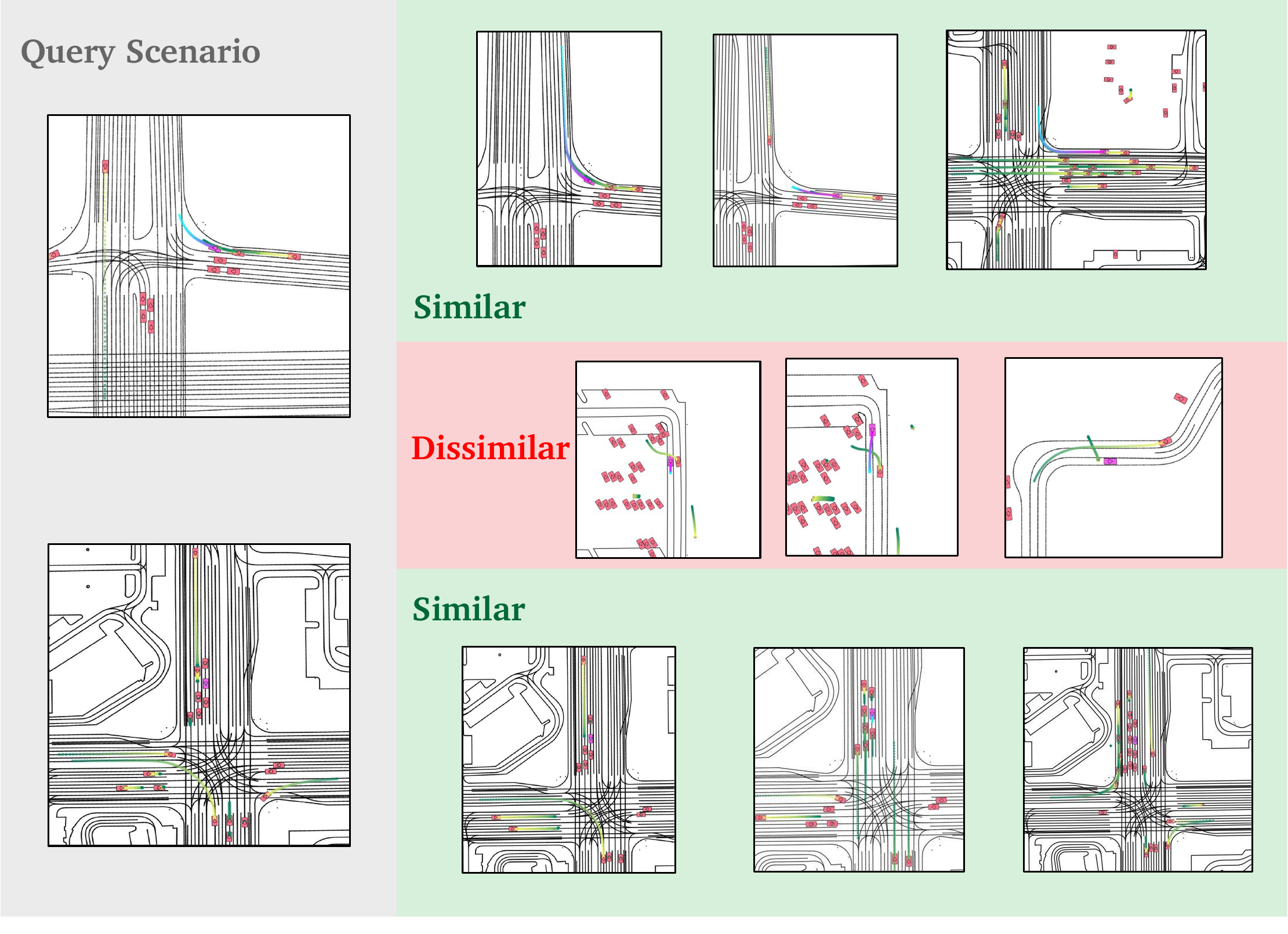}
    \caption{Visualization of retrieved scenarios. The left column shows the query rollout segment, while the right column displays the top semantically retrieved reference anchors.}
    \label{fig:retrieval}
    \vspace{-5mm}
\end{figure}

We provide visual examples of retrieved scenarios in Fig.~\ref{fig:retrieval}. The retrieved scenarios closely mirror the structural and dynamic topology of the query rollouts, further proving the robust faithfulness of our retrieval model. Importantly, the retrieved anchors accurately capture critical macro-level properties such as regional traffic density and complex road network geometries, fully validating the contextual reliability of our proposed RTE framework.

\subsection{Visualization of Long-term Simulation}
We provide additional visualizations of long-term simulation rollouts in Following figures. Each two row corresponds to a distinct scenario, with the top and bottom halves comparing InfGen and \methodname{}, respectively. The columns represent progressive time steps, illustrating how each model captures the evolving traffic dynamics. Notably, \methodname{} consistently generates more realistic traffic flows, with smoother agent trajectories and more natural interactions, while InfGen often produces abrupt insertions and unrealistic traffic patterns as time progresses. Please refer to the supplementary video for a more comprehensive visualization of these long-term simulations.
\newcommand{\tbox}[1]{\makebox[0.17\linewidth][c]{#1}}
\newcommand{\addFig}[1]{\includegraphics[width=1\linewidth]{figs/supp/vis/#1.jpg}}
\begin{figure*}
  \centering
  \footnotesize
  \setlength\tabcolsep{0mm}
  \begin{tabular}{cc cccccc}
      & & \tbox{$t=1$\,s} & \tbox{$t=6$\,s} & \tbox{$t=12$\,s} & \tbox{$t=18$\,s} & \tbox{$t=24$\,s} & \tbox{$t=30$\,s} \\
    & \rotatebox{90}{~~~~InfGen}\hspace{0mm} & \multicolumn{6}{c}{\addFig{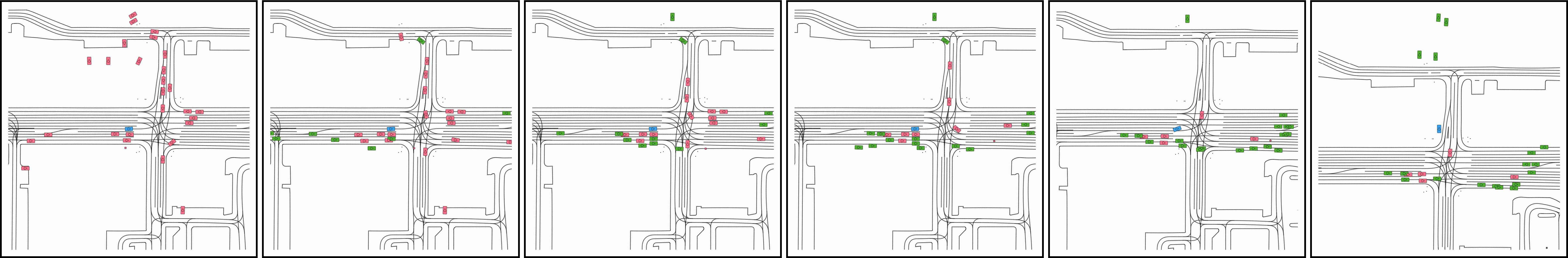}} \\
    & \rotatebox{90}{~\methodname{}}\hspace{0mm} & \multicolumn{6}{c}{\addFig{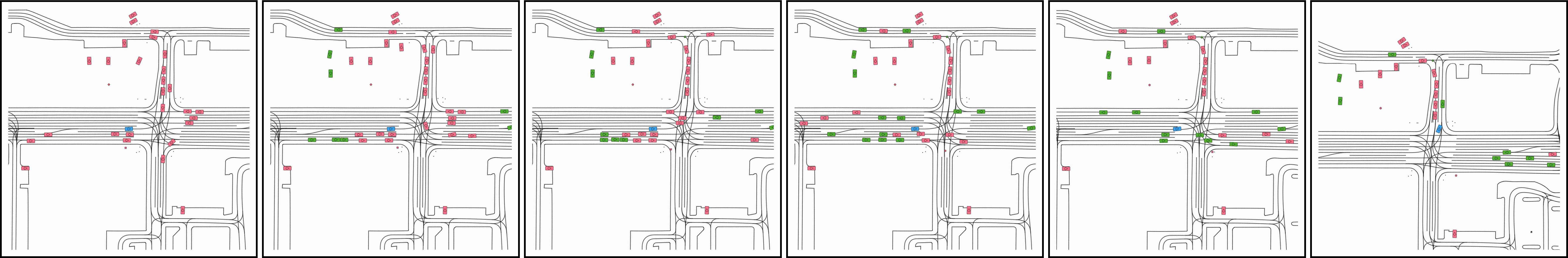}} \\
    & \rotatebox{90}{~~~~InfGen}\hspace{0mm} & \multicolumn{6}{c}{\addFig{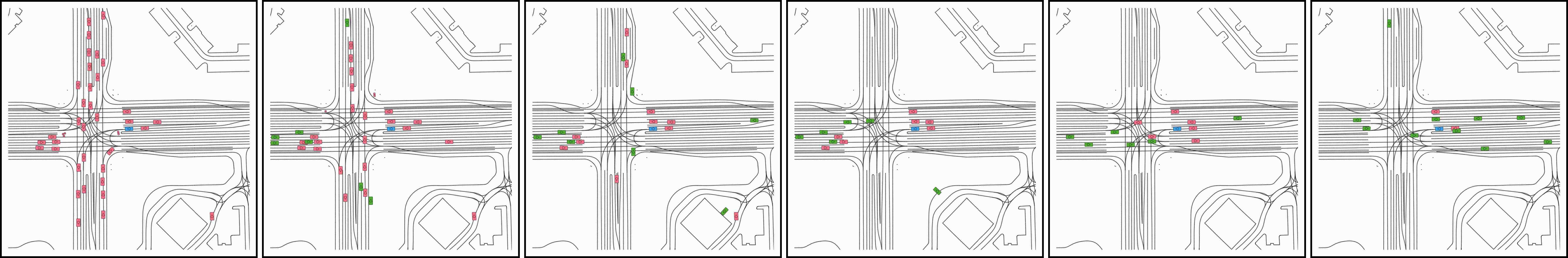}} \\
    & \rotatebox{90}{~\methodname{}}\hspace{0mm} & \multicolumn{6}{c}{\addFig{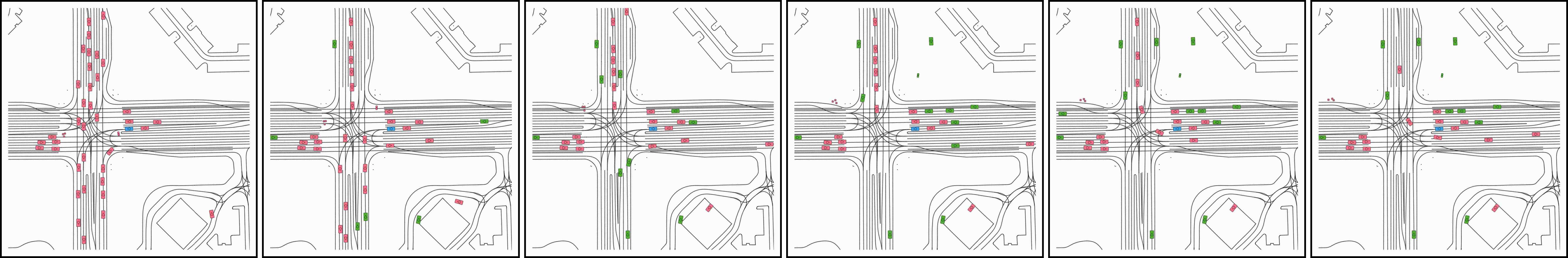}} \\
    & \rotatebox{90}{~~~~InfGen}\hspace{0mm} & \multicolumn{6}{c}{\addFig{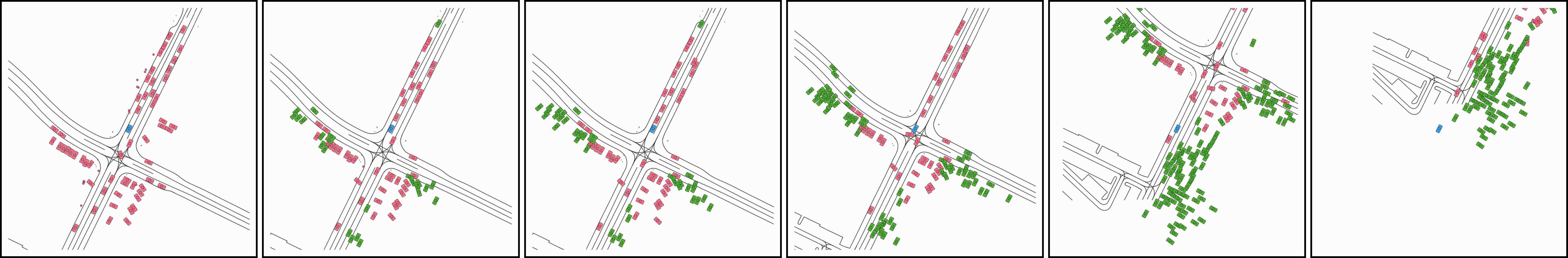}} \\
    & \rotatebox{90}{~\methodname{}}\hspace{0mm} & \multicolumn{6}{c}{\addFig{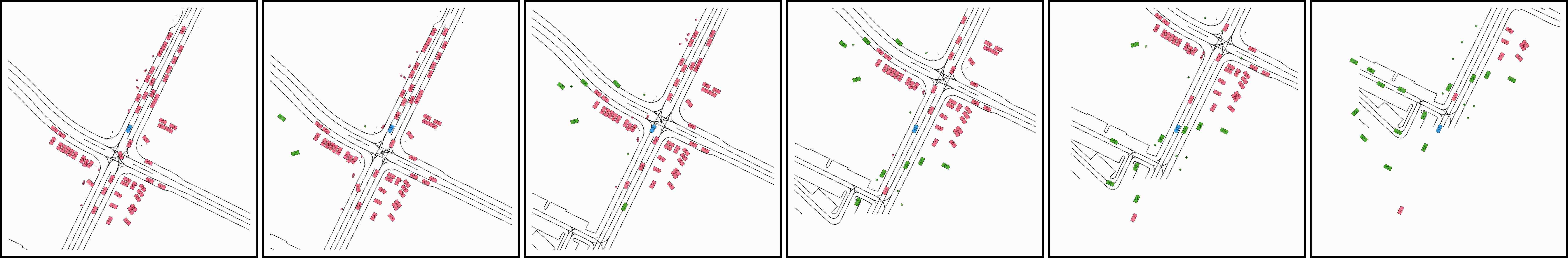}} \\
      & \rotatebox{90}{~~~~InfGen}\hspace{0mm} & \multicolumn{6}{c}{\addFig{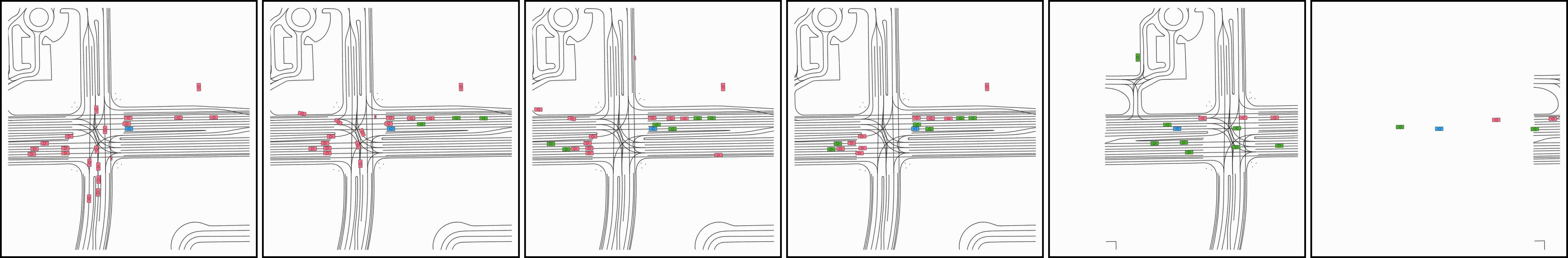}} \\
    & \rotatebox{90}{~\methodname{}}\hspace{0mm} & \multicolumn{6}{c}{\addFig{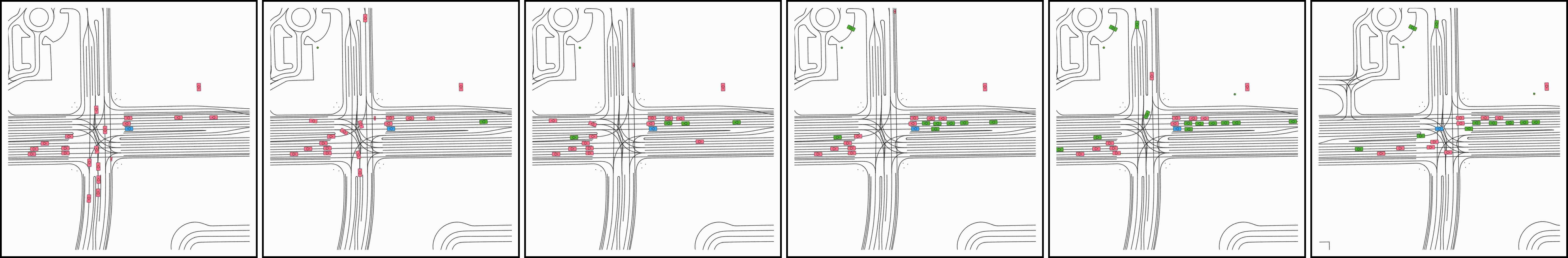}} \\
  \end{tabular}
  \caption{Visualization \#1}
\end{figure*}

\begin{figure*}
  \centering
  \footnotesize
  \setlength\tabcolsep{0mm}
  \begin{tabular}{cc cccccc}
      & & \tbox{$t=1$\,s} & \tbox{$t=6$\,s} & \tbox{$t=12$\,s} & \tbox{$t=18$\,s} & \tbox{$t=24$\,s} & \tbox{$t=30$\,s} \\
    & \rotatebox{90}{~~~~InfGen}\hspace{0mm} & \multicolumn{6}{c}{\addFig{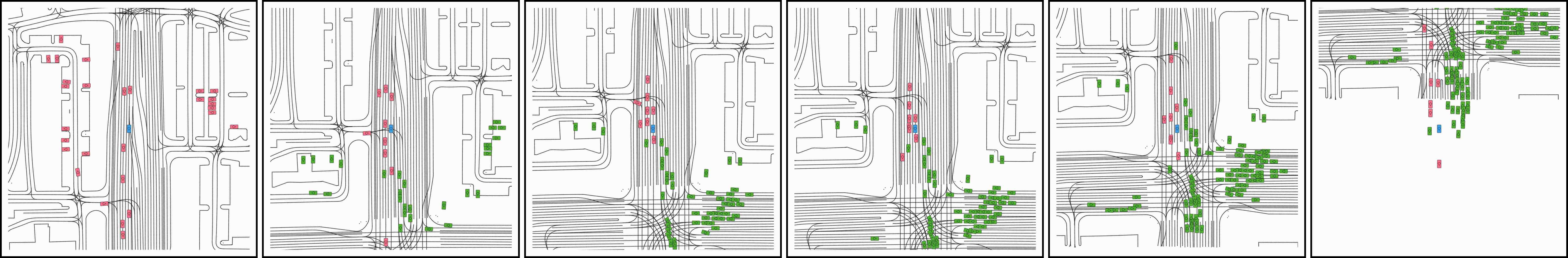}} \\
    & \rotatebox{90}{~\methodname{}}\hspace{0mm} & \multicolumn{6}{c}{\addFig{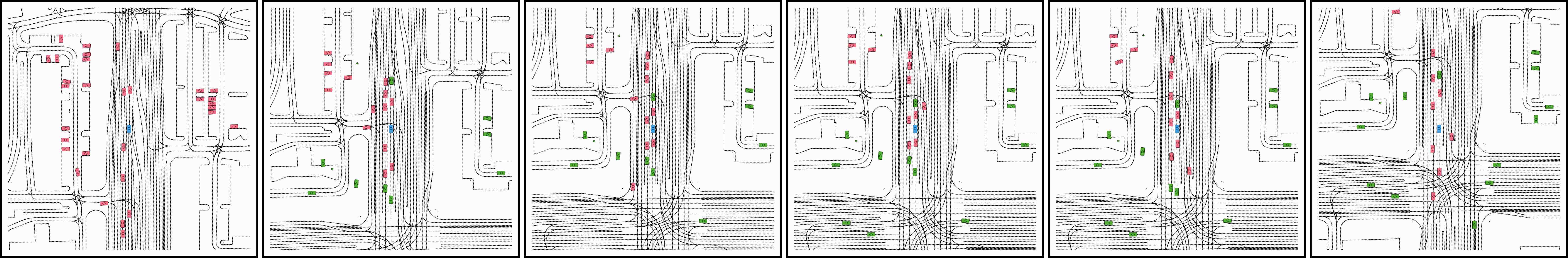}} \\
    & \rotatebox{90}{~~~~InfGen}\hspace{0mm} & \multicolumn{6}{c}{\addFig{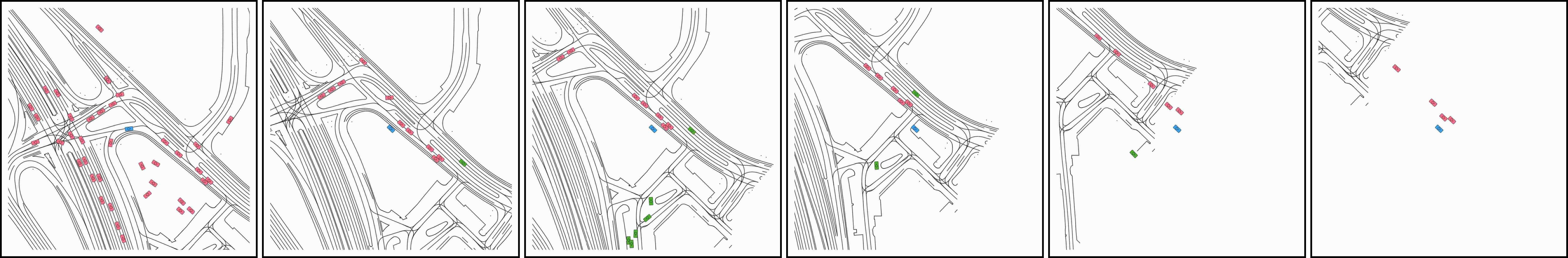}} \\
    & \rotatebox{90}{~\methodname{}}\hspace{0mm} & \multicolumn{6}{c}{\addFig{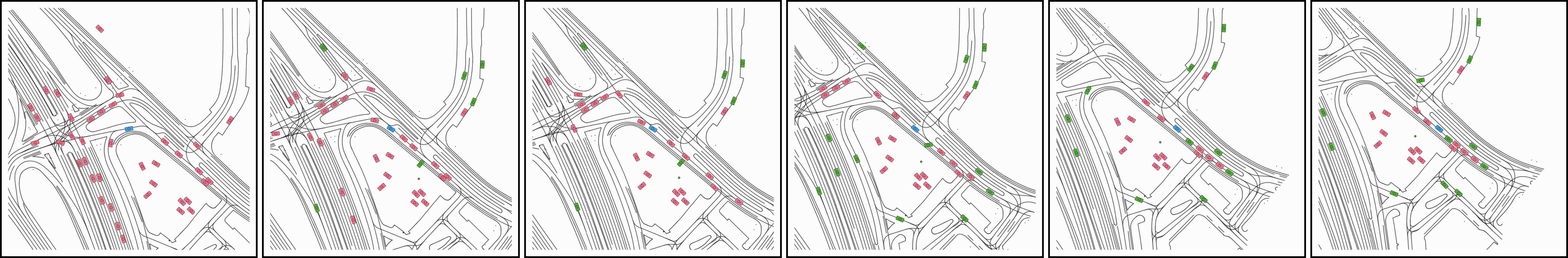}} \\
    & \rotatebox{90}{~~~~InfGen}\hspace{0mm} & \multicolumn{6}{c}{\addFig{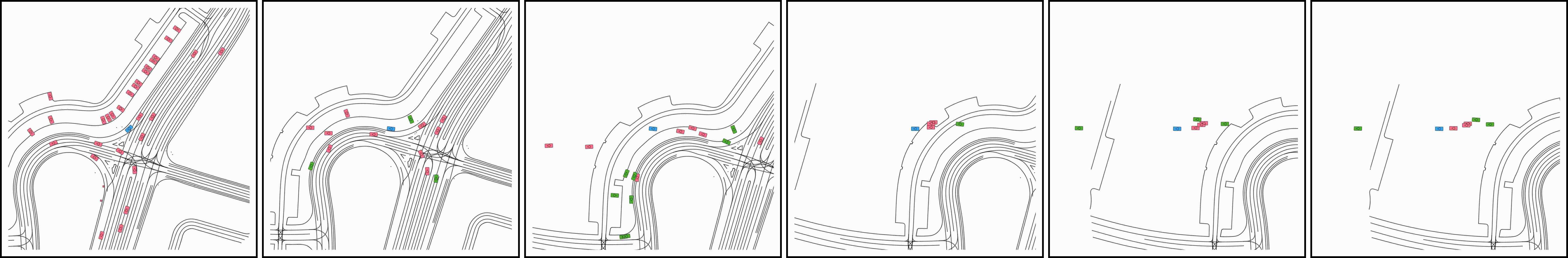}} \\
    & \rotatebox{90}{~\methodname{}}\hspace{0mm} & \multicolumn{6}{c}{\addFig{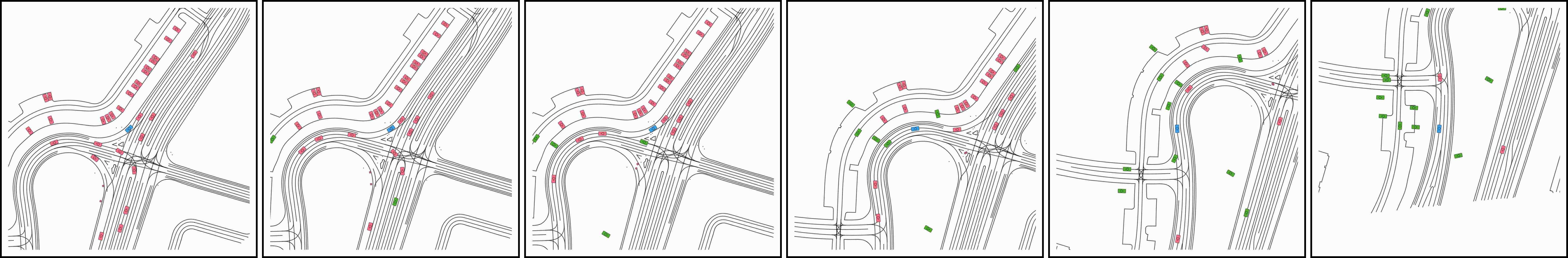}} \\
      & \rotatebox{90}{~~~~InfGen}\hspace{0mm} & \multicolumn{6}{c}{\addFig{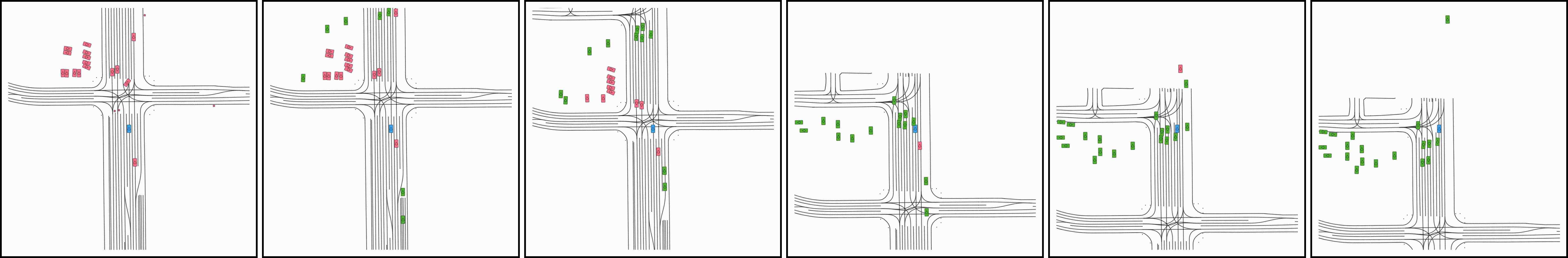}} \\
    & \rotatebox{90}{~\methodname{}}\hspace{0mm} & \multicolumn{6}{c}{\addFig{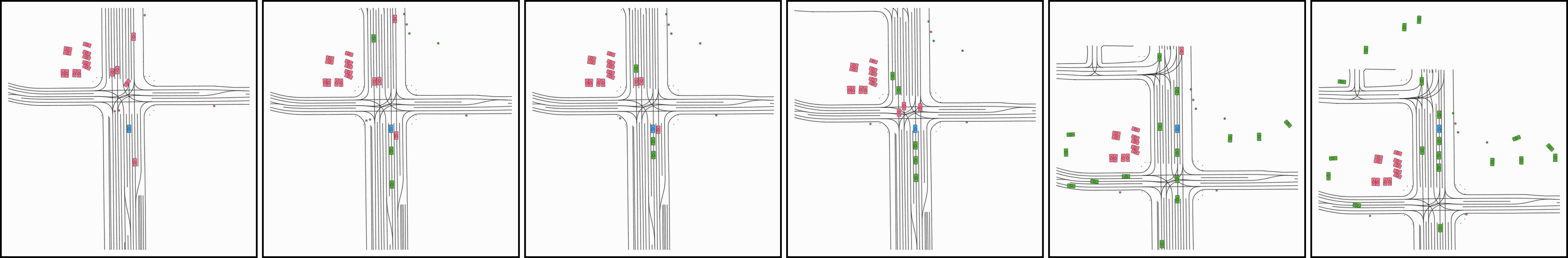}} \\
  \end{tabular}
  \caption{Visualization \#2}
\end{figure*}

\begin{figure*}
  \centering
  \footnotesize
  \setlength\tabcolsep{0mm}
  \begin{tabular}{cc cccccc}
      & & \tbox{$t=1$\,s} & \tbox{$t=6$\,s} & \tbox{$t=12$\,s} & \tbox{$t=18$\,s} & \tbox{$t=24$\,s} & \tbox{$t=30$\,s} \\
    & \rotatebox{90}{~~~~InfGen}\hspace{0mm} & \multicolumn{6}{c}{\addFig{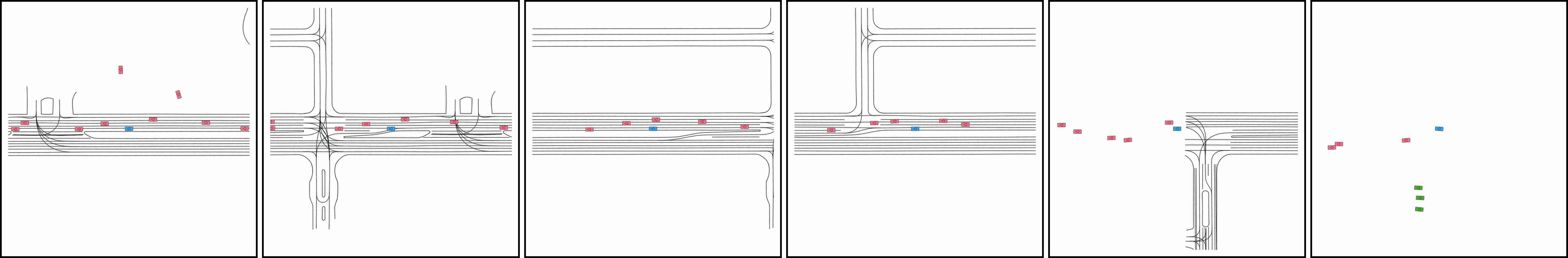}} \\
    & \rotatebox{90}{~\methodname{}}\hspace{0mm} & \multicolumn{6}{c}{\addFig{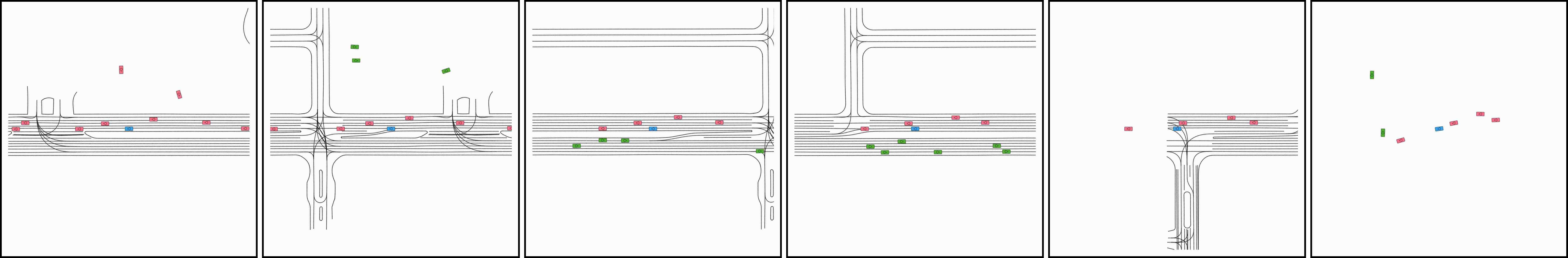}} \\
    & \rotatebox{90}{~~~~InfGen}\hspace{0mm} & \multicolumn{6}{c}{\addFig{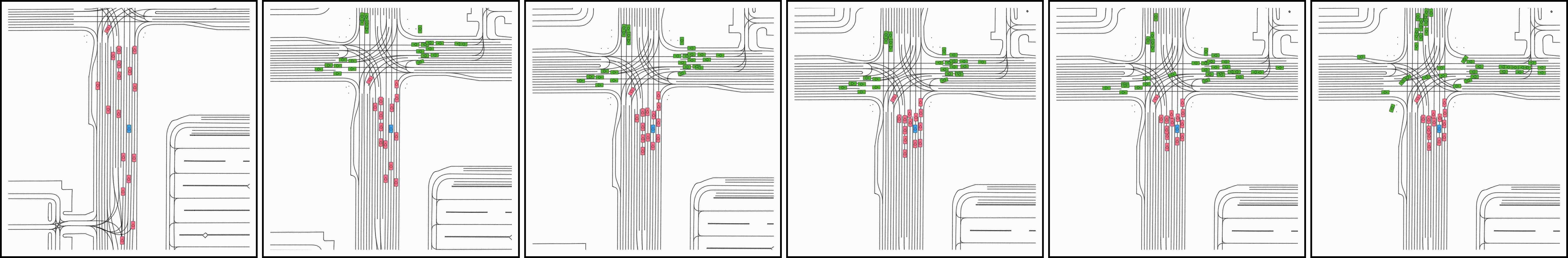}} \\
    & \rotatebox{90}{~\methodname{}}\hspace{0mm} & \multicolumn{6}{c}{\addFig{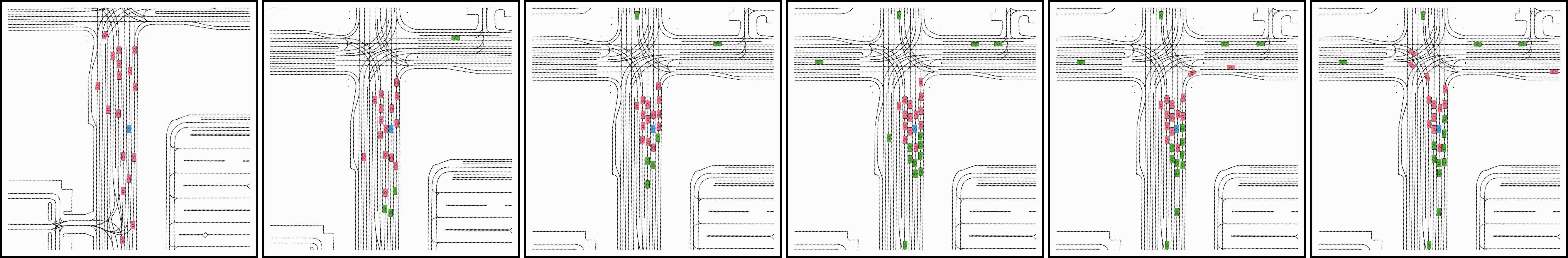}} \\
    & \rotatebox{90}{~~~~InfGen}\hspace{0mm} & \multicolumn{6}{c}{\addFig{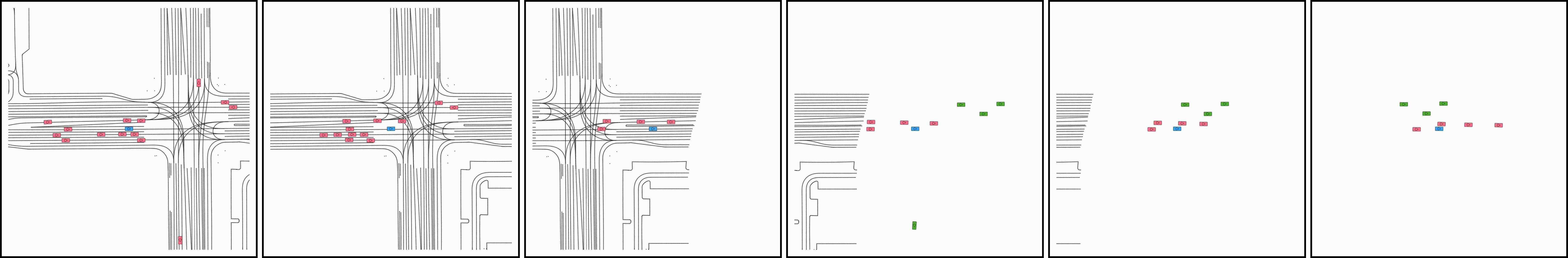}} \\
    & \rotatebox{90}{~\methodname{}}\hspace{0mm} & \multicolumn{6}{c}{\addFig{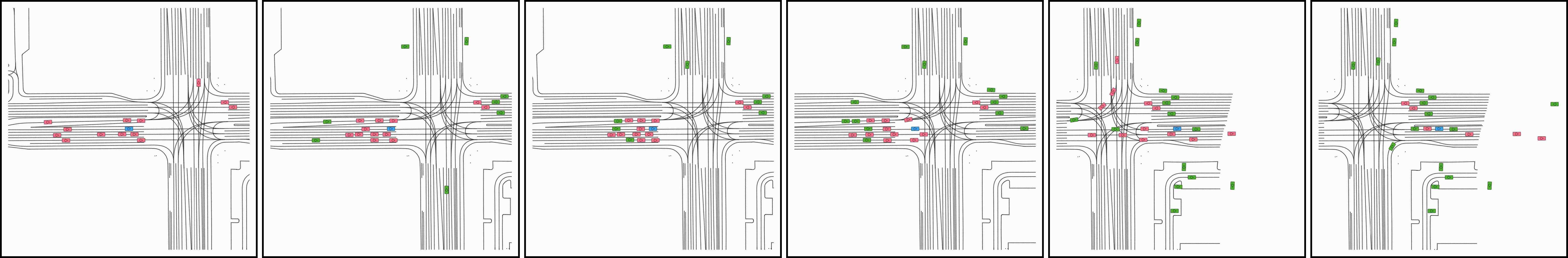}} \\
      & \rotatebox{90}{~~~~InfGen}\hspace{0mm} & \multicolumn{6}{c}{\addFig{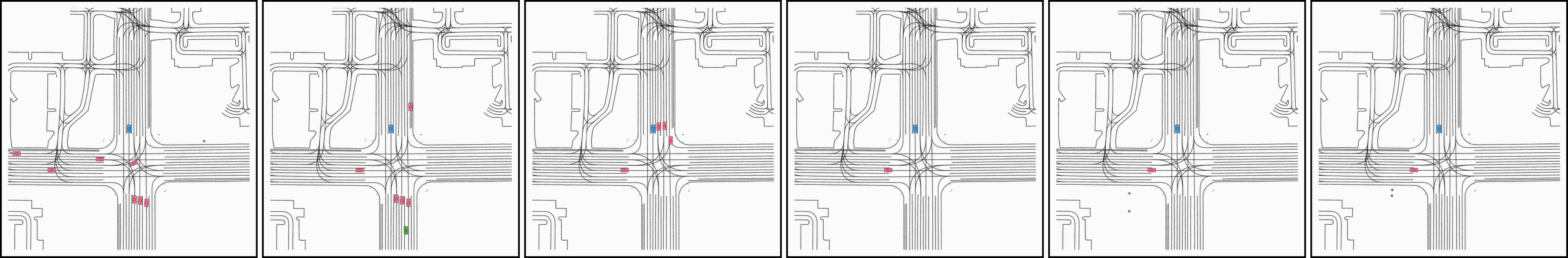}} \\
    & \rotatebox{90}{~\methodname{}}\hspace{0mm} & \multicolumn{6}{c}{\addFig{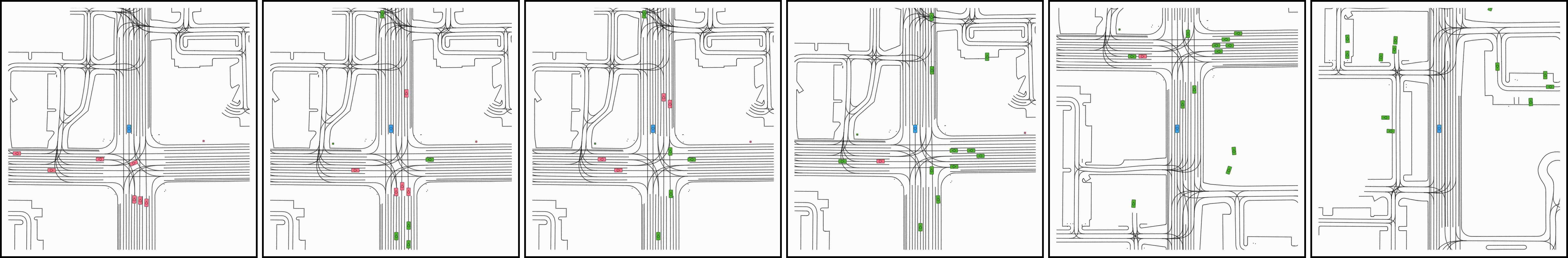}} \\
  \end{tabular}
  \caption{Visualization \#3}
\end{figure*}

\begin{figure*}
  \centering
  \footnotesize
  \setlength\tabcolsep{0mm}
  \begin{tabular}{cc cccccc}
      & & \tbox{$t=1$\,s} & \tbox{$t=6$\,s} & \tbox{$t=12$\,s} & \tbox{$t=18$\,s} & \tbox{$t=24$\,s} & \tbox{$t=30$\,s} \\
    & \rotatebox{90}{~~~~InfGen}\hspace{0mm} & \multicolumn{6}{c}{\addFig{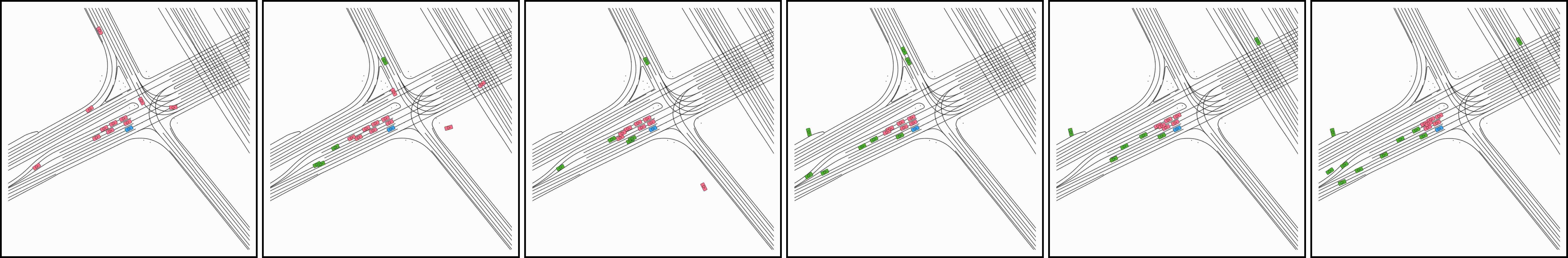}} \\
    & \rotatebox{90}{~\methodname{}}\hspace{0mm} & \multicolumn{6}{c}{\addFig{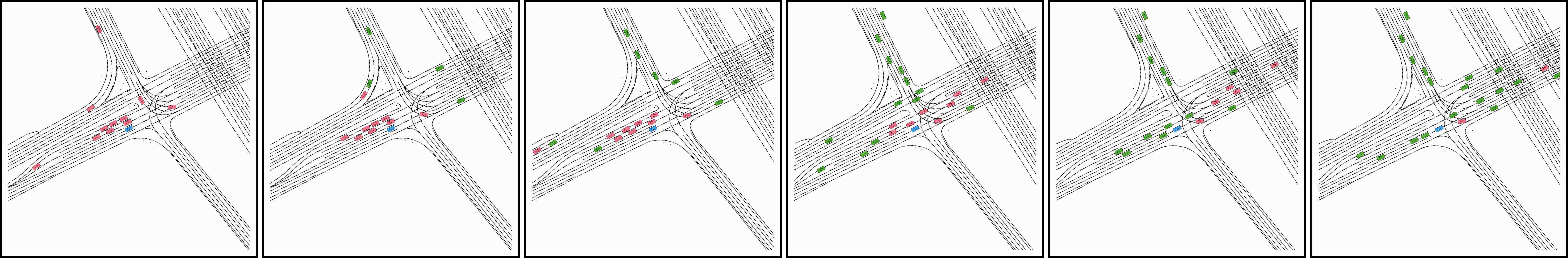}} \\
    & \rotatebox{90}{~~~~InfGen}\hspace{0mm} & \multicolumn{6}{c}{\addFig{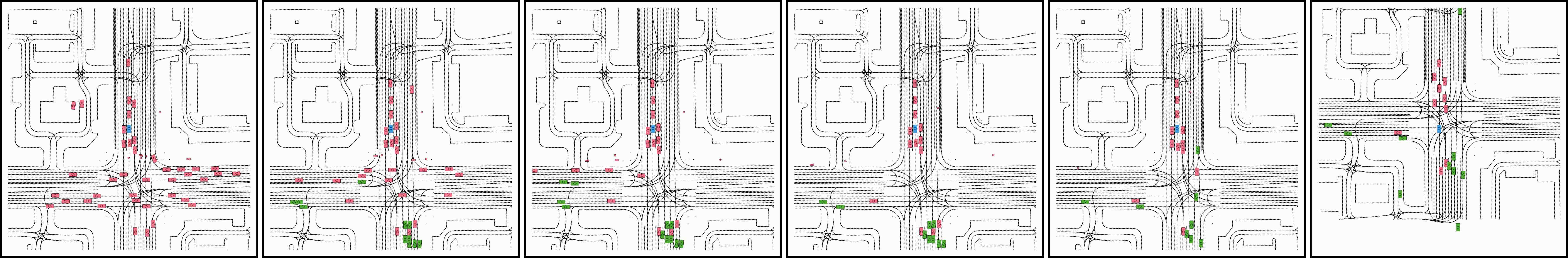}} \\
    & \rotatebox{90}{~\methodname{}}\hspace{0mm} & \multicolumn{6}{c}{\addFig{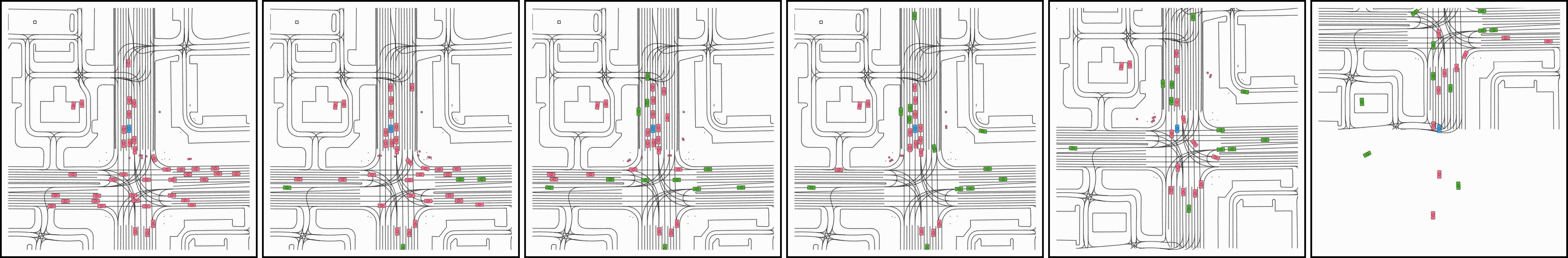}} \\
    & \rotatebox{90}{~~~~InfGen}\hspace{0mm} & \multicolumn{6}{c}{\addFig{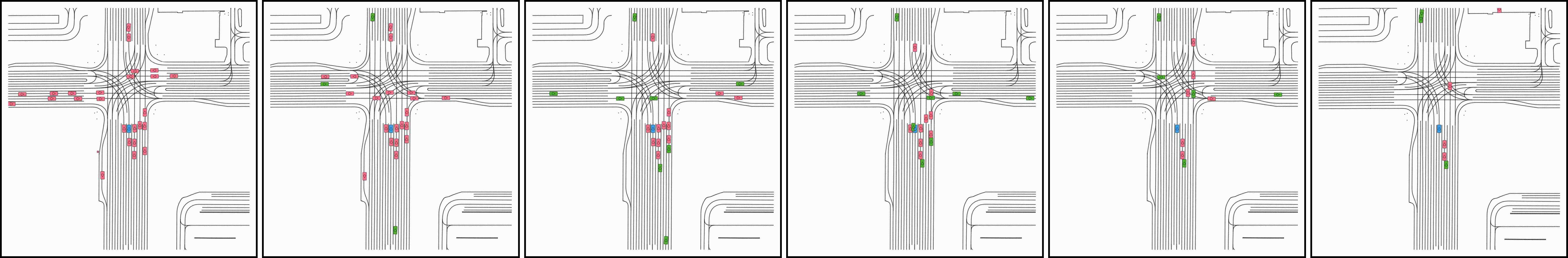}} \\
    & \rotatebox{90}{~\methodname{}}\hspace{0mm} & \multicolumn{6}{c}{\addFig{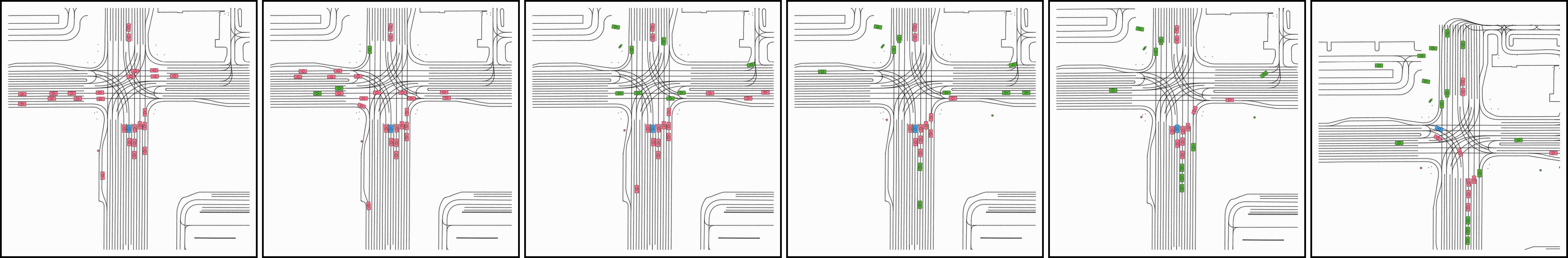}} \\
      & \rotatebox{90}{~~~~InfGen}\hspace{0mm} & \multicolumn{6}{c}{\addFig{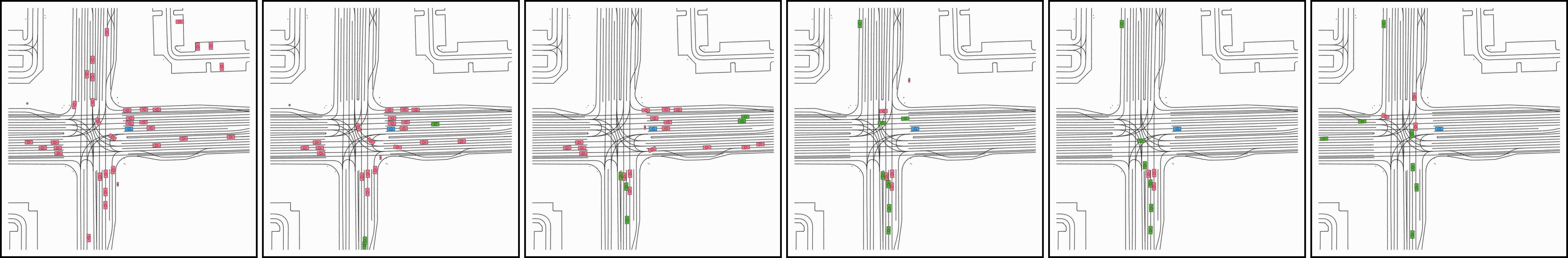}} \\
    & \rotatebox{90}{~\methodname{}}\hspace{0mm} & \multicolumn{6}{c}{\addFig{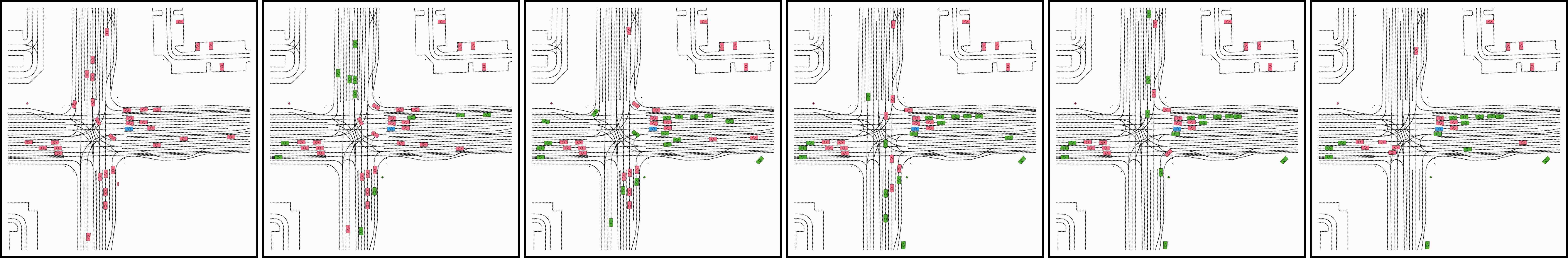}} \\
  \end{tabular}
  \caption{Visualization \#4}
\end{figure*}

\newpage
\section{Proofs and Pseudo-codes}
\subsection{Calculation of Attention Locality}

\begin{algorithm}[H]
\caption{Calculate Attention Locality}
\label{alg:locality_vec}
\begin{algorithmic}[1]
\REQUIRE Attention probability matrix $A \in \mathbb{R}^{L \times L}$, where $L$ is sequence length
\ENSURE Attention Locality Score $S_{loc} \in [0, 1]$
\STATE Initialize Mask Matrix $M \in \{0, 1\}^{L \times L}$ with zeros
\FOR{$i = 1$ to $L$}
    \FOR{$j = 1$ to $L$}
        \IF{$|i - j| \le 1$}
            \STATE $M_{i,j} \gets 1$ \COMMENT{Keep main diagonal and immediately adjacent diagonals}
        \ENDIF
    \ENDFOR
\ENDFOR
\STATE $A_{local} \gets A \odot M$ \COMMENT{Element-wise product to mask out distant attention}
\STATE $S_{token} \gets \sum_{j=1}^{L} A_{local}[:, j]$ \COMMENT{Sum local weights for each query token}
\STATE $S_{loc} \gets \text{Mean}(S_{token})$ \COMMENT{Average over the sequence length}
\RETURN $S_{loc}$
\end{algorithmic}
\end{algorithm}

\subsection{Proof: The Failure of NLL-Likelihood on \#Enter and \#Exit}
\label{sec.supp-placement-math}
Given the validation dataset, let the ground truth global spatial distribution for vehicle placement over 5 distance bins be empirically defined as:
\begin{equation}\nonumber
P_{GT} = [0.6852, 0.2280, 0.0623, 0.0175, 0.0069]
\end{equation}

Consider a non-generative baseline model (e.g., SMART or CAT-K) that is architecturally incapable of dynamically placing new vehicles. This model exhibits complete mode collapse, constantly outputting the most frequent boundary category (bin 0). Its predicted distribution for any given scenario effectively becomes:
\begin{equation}\nonumber
P_{model} = [1.0, 0.0, 0.0, 0.0, 0.0]
\end{equation}

\subsubsection{The NLL Paradox}
If we evaluate this collapsed model using pointwise Negative Log-Likelihood (NLL) against the global distribution, the model safely selects index 0 every time. The resulting NLL is low and constant:
\begin{equation}\nonumber
NLL_{model} = -\log(P_{GT}[0]) = -\log(0.6852) \approx 0.378
\end{equation}

To expose why this is mathematically problematic, we compare it against a theoretical ``perfect'' generative model that samples precisely according to the true distribution $P_{GT}$. The expected NLL for this perfect model is equivalent to the Shannon Entropy of $P_{GT}$:
\begin{equation}\nonumber
\begin{split}
\mathbb{E}[NLL_{perfect}] &= H(P_{GT}) = -\sum_{i=1}^{5} P_{GT}[i] \log(P_{GT}[i]) \\
&= -(0.6852 \log 0.6852 + \dots + 0.0069 \log 0.0069) \approx 0.993
\end{split}
\end{equation}

Here lies the paradox: $NLL_{model} (0.378) \ll \mathbb{E}[NLL_{perfect}] (0.993)$. The NLL metric perversely rewards the collapsed heuristic model for safely hugging the highest-probability mode, while severely penalizing the perfect generative model for correctly attempting to simulate rare, long-tail events (which incur massive pointwise NLL penalties, e.g., $-\log(0.0069) \approx 4.97$). Therefore, pointwise NLL is fundamentally incapable of evaluating generative diversity and fidelity.

\begin{algorithm}[ht]
\caption{Calculation of Distributional Divergence $D$ (Placement)}
\label{alg:calc_d}
\begin{algorithmic}[1]
\REQUIRE Simulated values $V_{sim} \in \mathbb{R}^{N}$, Ground truth probabilities $P_{GT} \in \mathbb{R}^{K}$
\REQUIRE Config: $min = 0.0$, $max = 10.0$, $K = 5$, $\alpha = 0.1$
\STATE \textbf{Initialize:} Bin edges $E \leftarrow \text{linspace}(min, max, K+1)$
\STATE \textbf{Initialize:} Counts vector $C \leftarrow \mathbf{0} \in \mathbb{R}^{K}$
\FOR{each $v \in V_{sim}$}
    \STATE Find bin index $k$ such that $E_k \le v < E_{k+1}$
    \STATE $k \leftarrow \text{clamp}(k, 0, K-1)$
    \STATE $C_k \leftarrow C_k + 1$
\ENDFOR
\STATE \textbf{Apply Additive Smoothing:} $C \leftarrow C + \alpha$
\STATE \textbf{Normalize to Probabilities:} $P_{sim} \leftarrow \frac{C}{\sum_{i=1}^K C_i}$
\STATE \textbf{Compute Mean Distribution:} $M \leftarrow \frac{1}{2}(P_{sim} + P_{GT})$
\STATE \textbf{Compute KL Divergences:}
\STATE $\quad D_{KL}(P_{sim} \parallel M) \leftarrow \sum_{k=1}^K P_{sim, k} \log\left(\frac{P_{sim, k}}{M_k}\right)$
\STATE $\quad D_{KL}(P_{GT} \parallel M) \leftarrow \sum_{k=1}^K P_{GT, k} \log\left(\frac{P_{GT, k}}{M_k}\right)$
\STATE \textbf{Compute JS Divergence:} $D_{JS} \leftarrow \frac{1}{2} \left( D_{KL}(P_{sim} \parallel M) + D_{KL}(P_{GT} \parallel M) \right)$
\STATE \textbf{Calculate Final Score:} $D\_Score \leftarrow \max\left(0, 1 - \frac{D_{JS}}{\log 2}\right)$
\RETURN $D\_Score$
\end{algorithmic}
\end{algorithm}

\subsubsection{Resolution via Jensen-Shannon Divergence}
To accurately evaluate true distributional alignment, we enforce the Jensen-Shannon (JS) Divergence. We compute $D_{JS}(P_{model} \parallel P_{GT})$ based on the mean distribution $M$:
\begin{equation}\nonumber
M = \frac{1}{2}(P_{model} + P_{GT}) = [0.8426, 0.1140, 0.03115, 0.00875, 0.00345]
\end{equation}

The KL divergence from the collapsed model to $M$ is:
\begin{equation}\nonumber
D_{KL}(P_{model} \parallel M) = 1.0 \times \log\left(\frac{1.0}{0.8426}\right) + 0 \approx 0.1712
\end{equation}

The KL divergence from $P_{GT}$ to $M$ is:
\begin{equation}\nonumber
D_{KL}(P_{GT} \parallel M) = \sum_{i=1}^{5} P_{GT}[i] \log\left(\frac{P_{GT}[i]}{M[i]}\right)
\end{equation}
Given $P_{GT}[i] = 2 M[i]$ for $i \in \{2,3,4,5\}$, we have:
\begin{equation}\nonumber
\begin{split}
D_{KL}(P_{GT} \parallel M) &= 0.6852 \log\left(\frac{0.6852}{0.8426}\right) + \sum_{i=2}^{5} P_{GT}[i] \log(2) \\
&\approx -0.1417 + 0.3148 \times 0.6931 \approx 0.0764
\end{split}
\end{equation}

The final JS Divergence cleanly resolves to:
\begin{equation}\nonumber
D_{JS} = \frac{1}{2}(0.1712 + 0.0764) = 0.1238
\end{equation}

Unlike NLL, the JS Divergence strictly bounds and penalizes the distance between the collapsed output distribution and the true generative distribution. A $D_{JS}$ of $0.1238$ maps directly to a significant penalty in the final normalized score (e.g., $1 - \frac{0.1238}{\log 2} \approx 0.82$), properly punishing the baseline for its inability to model long-tail behaviors. The pseudo code for this calculation is provided in Algorithm~\ref{alg:calc_d}.

\subsection{Proof: Problem on $D_{enter}$ and $D_{exit}$ Metrics}
\label{sec:metric_exploitation}

Here, we provide a formal mathematical proof demonstrating why evaluating enter distance ($D_{enter}$) and exit distance ($D_{exit}$) via pointwise NLL is fundamentally flawed. We prove that these metrics inherently reward non-generative, heuristic policies while heavily penalizing architectures that accurately generate real traffic events.

\subsubsection{Empirical Distribution and the Metric Definition}
In real-world long-term scenarios, vehicle insertions and deletions are spatially skewed toward the absolute boundary of the sensor range. Taking the ground truth placement distance distribution $P_{GT}$ over 10 bins as an example, empirical validation yields:
\begin{equation}\nonumber
P_{GT} = [0.0013, 0.0078, \dots, 0.1701, \mathbf{0.3371}]
\end{equation}
The probability mass is heavily concentrated in the outermost distance bin ($p_9 = 0.3371$).

The evaluation score for a simulated distribution $P_{sim}$ is calculated as the geometric mean of the likelihoods, equivalent to the exponential of the expected negative log-likelihood:
\begin{equation}\nonumber
\text{Score} = \exp \left( \mathbb{E}_{x \sim P_{sim}} [\ln P_{GT}(x)] \right)
\end{equation}

\subsubsection{Heuristic Exploitation by Non-Generative Models}
Short-term optimized models (e.g., SMART, CAT-K) lack the ability to dynamically insert new vehicles mid-scene. In the standard evaluation implementations, the absence of a localized placement event is heuristically assigned a maximum distance value (forced to the perception boundary).

As a result, the predictions of these non-generative models overwhelmingly collapse into the final bin. Assuming the ideal heuristic case where all placements are mapped to the outermost bin, the model's output distribution becomes a one-hot vector:
\begin{equation}\nonumber
P_{heuristic} = [0, 0, \dots, 0, 1.0]
\end{equation}

When evaluated, this heuristic policy constantly hits the highest probability bin ($p_9$). Its expected score is strictly bounded by the maximum pointwise likelihood:
\begin{equation}\nonumber
\text{Score}_{heuristic} = \exp \left( \ln P_{GT}[9] \right) = P_{GT}[9] = \mathbf{0.3371}
\end{equation}
\textit{(Note: In practice, due to early-step initializations, a negligible fraction might fall into adjacent bins, resulting in an empirical score slightly less than but asymptotically approaching $0.3371$. This perfectly aligns with the officially reported $0.3371$ for both CAT-K and SMART.)}

\subsubsection{The Penalty on Perfect Generation}
Now, consider a theoretically perfect generative model that successfully captures true spatial diversity. Its generated samples perfectly match the ground truth distribution: $P_{perfect} \equiv P_{GT}$.

The expected score for this perfect model is strictly governed by the Shannon Entropy $H(P_{GT})$ of the ground truth distribution:
\begin{equation}\nonumber
\begin{split}
\text{Score}_{perfect} &= \exp \left( \sum_{i=0}^{9} P_{GT}[i] \ln(P_{GT}[i]) \right) = \exp \left( -H(P_{GT}) \right)
\end{split}
\end{equation}

Using the empirical $P_{GT}$ values to compute the entropy:
\begin{equation}\nonumber
\begin{split}
-H(P_{GT}) &= 0.0013 \ln(0.0013) + \dots + 0.3371 \ln(0.3371) \\
&\approx -1.849 \text{ nats}
\end{split}
\end{equation}

The theoretical upper bound for the perfect generative model evaluates to:
\begin{equation}\nonumber
\text{Score}_{perfect} = \exp(-1.849) \approx \mathbf{0.157}
\end{equation}

\subsubsection{Conclusion}
The mathematical proof reveals a critical scoring paradox:
\begin{equation}\nonumber
\text{Score}_{perfect} (0.157) \ll \text{Score}_{heuristic} (0.3371)
\end{equation}

The pointwise likelihood metric structurally penalizes the generation of diverse, long-tail spatial events (e.g., vehicles merging realistically from occluded driveways). Instead, it heavily rewards mode-seeking behavior, allowing models with zero generative capabilities to mathematically ``hack'' the metric by defaulting to boundary states.

Given that $D_{enter}$ and $D_{exit}$ are inherently adversarial to the goal of simulating diverse and realistic traffic, and are trivially exploitable by static baselines, they provide no meaningful signal regarding true generative fidelity. Therefore, we explicitly omit the $D_{enter}$ and $D_{exit}$ metrics from our comprehensive evaluation protocol.

\end{document}